\documentclass[letterpaper]{article} 
\usepackage{aaai25}  
\usepackage{times}  
\usepackage{helvet}  
\usepackage{courier}  
\usepackage[hyphens]{url}  
\usepackage{graphicx} 
\usepackage{natbib}  
\usepackage{caption} 
\usepackage{algorithm}
\usepackage{algorithmic}
\usepackage{amsmath} 
\usepackage{amssymb} 
\usepackage{multirow} 
\usepackage{booktabs} 
\usepackage{xcolor}
\usepackage{newfloat}
\usepackage{listings}
\DeclareCaptionStyle{ruled}{labelfont=normalfont,labelsep=colon,strut=off} 
\floatstyle{ruled}
\newfloat{listing}{tb}{lst}{}
\floatname{listing}{Listing}
\title{Diff-Shadow: Global-guided Diffusion Model for Shadow Removal}
\author{
    Jinting Luo\textsuperscript{\rm 1},
    Ru Li\textsuperscript{\rm 2},
    Chengzhi Jiang\textsuperscript{\rm 1},
    Xiaoming Zhang\textsuperscript{\rm 3},
    \\Mingyan Han\textsuperscript{\rm 1},
    Ting Jiang\textsuperscript{\rm 1},
    Haoqiang Fan\textsuperscript{\rm 1},
    Shuaicheng Liu\textsuperscript{\rm 4}\thanks{Corresponding author}
}
\affiliations{
    \textsuperscript{\rm 1}Megvii Technology Inc.
    \textsuperscript{\rm 2}Harbin Institute of Technology\\
    \textsuperscript{\rm 3}Southwest Jiaotong University
    \textsuperscript{\rm 4}University of Electronic Science and Technology of China\\

    \{k531756653, morven126, zxiaoming360, hanmy628, tjhedlen\}@gmail.com,\\ liru@hit.edu.cn,  fhq@megvii.com, liushuaicheng@uestc.edu.cn
}

\usepackage{bibentry}

\begin{document}

\maketitle

\begin{abstract}
We propose Diff-Shadow, a global-guided diffusion model for shadow removal. Previous transformer-based approaches can utilize global information to relate shadow and non-shadow regions but are limited in their synthesis ability and recover images with obvious boundaries. In contrast, diffusion-based methods can generate better content but they are not exempt from issues related to inconsistent illumination. In this work, we combine the advantages of diffusion models and global guidance to achieve shadow-free restoration. Specifically, we propose a parallel UNets architecture: 1) the local branch performs the patch-based noise estimation in the diffusion process, and 2) the global branch recovers the low-resolution shadow-free images. A Reweight Cross Attention (RCA) module is designed to integrate global contextual information of non-shadow regions into the local branch. We further design a Global-guided Sampling Strategy (GSS) that mitigates patch boundary issues and ensures consistent illumination across shaded and unshaded regions in the recovered image. Comprehensive experiments on datasets ISTD, ISTD+, and SRD have demonstrated the effectiveness of Diff-Shadow. Compared to state-of-the-art methods, our method achieves a significant improvement in terms of PSNR, increasing from 32.33dB to 33.69dB on the ISTD dataset. 
\begin{links}
    \link{Code}{https://github.com/MonteCarluo/Diff-Shadow}
\end{links}
\end{abstract}

\section{Introduction} \label{sec:intro}

Shadows are a phenomenon that exists when an optical image is captured with light blocked. The presence of shadow can complicate image processing as well as many subsequent vision tasks, e.g., object detection, tracking, and semantic segmentation~\cite{guo2023shadowformer}. The goal of shadow removal is to enhance the brightness of the image shadow regions and to achieve a consistent illumination distribution between shadow and non-shadow regions. The shadow removal task is essentially an image restoration task designed to utilize the information available in the shadow-affected image to recover the information lost due to light occlusion. 

\begin{figure}[t]
\setlength{\belowcaptionskip}{-0.4cm}
    \centering
    \includegraphics[width=1.0\linewidth]{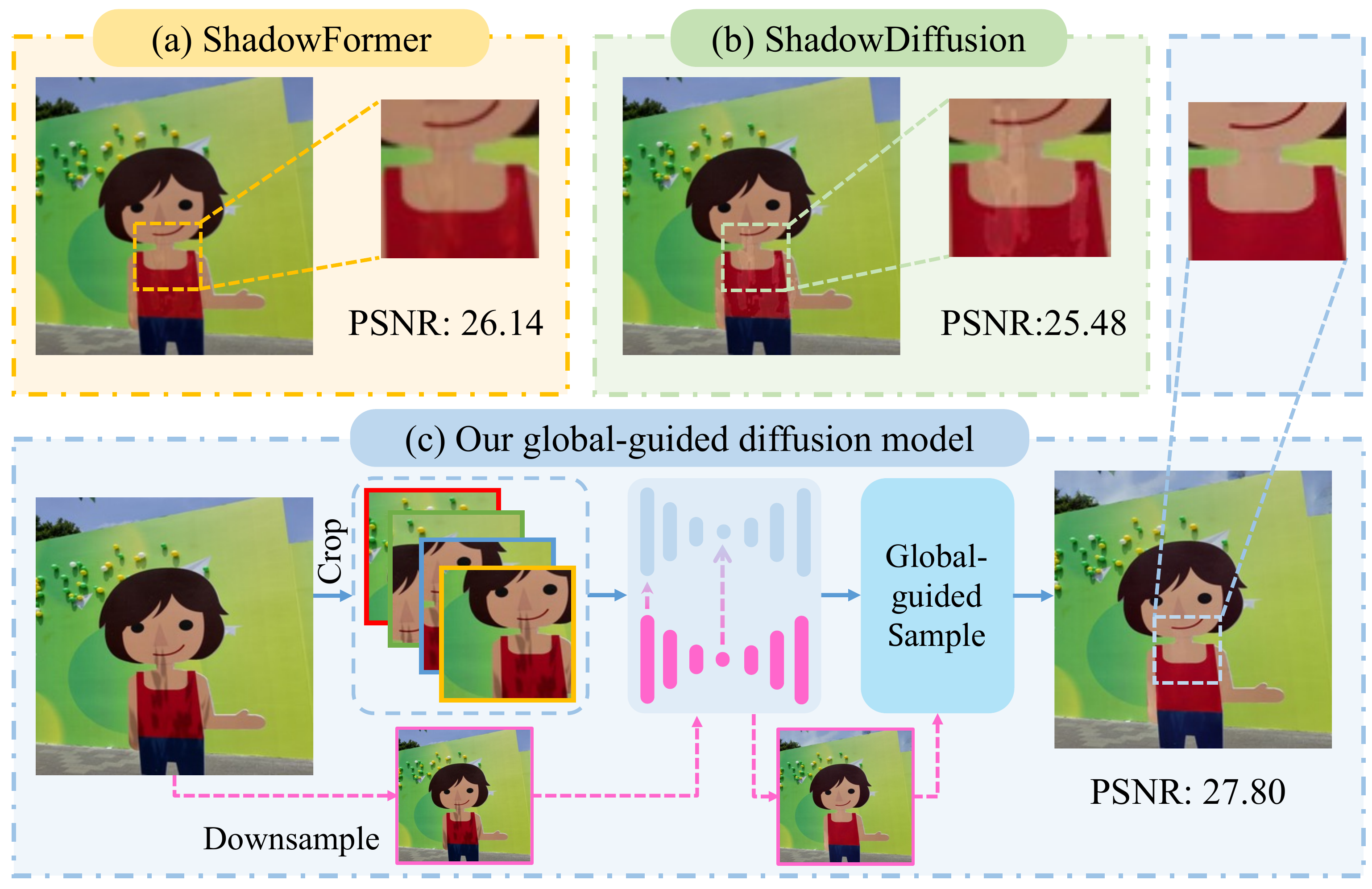}
    \caption{(a) shows the result of ShadowFormer~\cite{guo2023shadowformer}, which suffers from residual shadow artifacts due to the limited modeling ability, while the result of (b) ShadowDiffusion~\cite{guo2023shadowdiffusion} shows obvious illumination inconsistency across the images because it cannot exploit the global information. (c) exhibits the proposed Diff-Shadow, which 
    generates high-quality shadow removal results that maintain illumination consistency and are free from block boundary artifacts through the design of a parallel network structure and a novel global-guided sampling strategy.}
    \label{fig:teaser}
\end{figure}

Shadow removal has received intensive and long-lasting attention. Traditional approaches use hand-crafted features and are based on physical lighting models~\cite{gryka2015learning, zhang2015shadow, xiao2013fast}. Inspired by the successes of deep learning, many approaches that utilize large amounts of data~\cite{guo2023shadowformer, chen2021canet, fu2021auto, zhu2022bijective} have been introduced to improve shadow removal and reveal promising improvements, ~\emph{e.g.}, Guo~\emph{et al.} proposed a transformer-based architecture, named ShadowFormer, to exploit non-shadow
information to help shadow region restoration. Their shadow-interaction module plays an important role in exploring global information. However, 
this method often overlooks the modeling of the underlying distribution of natural images. Consequently, the generated results appear to have significant residual shadows, as shown in Fig.~\ref{fig:teaser} (a).

Recently, diffusion models have gained wide interest in the field of image restoration~\cite{saharia2022palette, saharia2022image, jiang2023low} due to its powerful modeling ability.
Guo~\emph{et al.} proposed ShadowDiffusion to jointly pursue shadow-free images and refined shadow masks~\cite{guo2023shadowdiffusion}. 
 
Although it covers the whole input image, the kind of conventional diffusion-based methods use a simple UNet and are challenging for shadow regions to capture effective information from matched non-shadow regions, resulting in illumination inconsistency in Fig.~\ref{fig:teaser} (b). 
Furthermore, generating arbitrary resolution images directly with ShadowDiffusion proved challenging as diffusion models are more suitable for processing smaller patches~\cite{ding2023patched}. While Zheng~\emph{et al.}~\cite{ding2023patched} and Ozan~\emph{et al.}~\cite{ozdenizci2023restoring} proposed several patch-based diffusion methods that demonstrated exceptional performance in image restoration tasks at arbitrary resolutions, they still faced limitations when handling shadow removal task. These methods lack global contextual information and ultimately lead to lighting inconsistencies or obvious boundaries in the restored images.

Considering the powerful modeling capabilities of diffusion models and the necessity of global information for shadow removal tasks,
we propose Diff-Shadow, a global-guided diffusion model for generating high-quality shadow-free results with no obvious boundaries and consistent illumination in arbitrary image resolution. Our Diff-Shadow adopts the parallel UNets network architecture that receives both patch and global inputs, where the local branch focuses on estimating patch noise, while the global branch is responsible for recovering low-resolution shadow-free images. Additionally, our Diff-Shadow exploits the global contextual information of the non-shadow region from the global branch through the Reweight Cross Attention (RCA) module.
Subsequently, we present a Global-guided Sampling Strategy (GSS) within the denoising process. In this strategy, the fusion weights for patch noise are determined by considering both the brightness disparity between the recovered patch and the global image, as well as the extent of shadow regions within the patch. This approach addresses potential merging artifacts, preserves superb illumination coherence across shaded and non-shadow regions, and ultimately enhances the overall quality of the results, as shown in Fig.~\ref{fig:teaser} (c). Moreover, benefiting from the patch-based methods, the Diff-Shadow can be expanded to handle images with arbitrary resolution.

To summarize, our main contributions are as follows: 

(1) We propose Diff-Shadow, a global-guided diffusion architecture that incorporates parallel UNets and exploits the correlation between the local branch and the non-shadow regions of the global through a reweighted cross-attention (RCA) module, which removes shadows efficiently.

(2) We propose a Global-guided Sampling Strategy (GSS) to further improve the quality of shadow removal during the diffusion denoising process.

(3) Extensive experimental results on public ISTD~\cite{wang2018stacked}, ISTD+~\cite{le2019shadow}, and SRD~\cite{qu2017deshadownet} datasets demonstrate that the proposed Diff-Shadow outperforms the state of the art.

\section{Related Works}
\label{sec:relatedworks}

\noindent\textbf{Shadow Removal.}
Recent learning-based methods for shadow removal tasks have achieved superior performance by leveraging large-scale high-quality training data to learn high-level image texture and semantic information~\cite{chen2021canet,fu2021auto,guo2023shadowformer,guo2023shadowdiffusion}. Unlike traditional physical-based methods that rely on prior assumptions like illumination~\cite{arbel2010shadow,zhang2015shadow}, gradient~\cite{finlayson2002removing,finlayson2005removal,mohan2007editing,liu2008texture,gryka2015learning}, and region~\cite{guo2012paired,guo2021self,guo2022exploiting}, the learning-based approaches tend to learn contextual features for shadow removal. For instance, Qu~\emph{et al.} proposed a multi-context embedding network to integrate multi-level information~\cite{qu2017deshadownet}, while Wang~\emph{et al.} explored the relationship between shadow detection and removal, and proposed a stacked conditional generative adversarial network (ST-CGAN)~\cite{wang2018stacked}. Zhu~\emph{et al.} proposed a model-driven network~\cite{zhu2022efficient} combining efficiency and interpretability by formulating a new shadow illumination model and unfolding it into a deep network. Other works like Hu~\emph{et al.}~\cite{hu2019direction} and Zhang~\emph{et al.}~\cite{zhang2020ris} focused on applying spatial context attention features and exploring residual-illumination relationships, respectively. These learning-based methods have produced impressive results, but there remain opportunities to further capture the correspondence between shadow and non-shadow areas.

\begin{figure*}[t]
    \centering
    \includegraphics[width=1.0\linewidth]{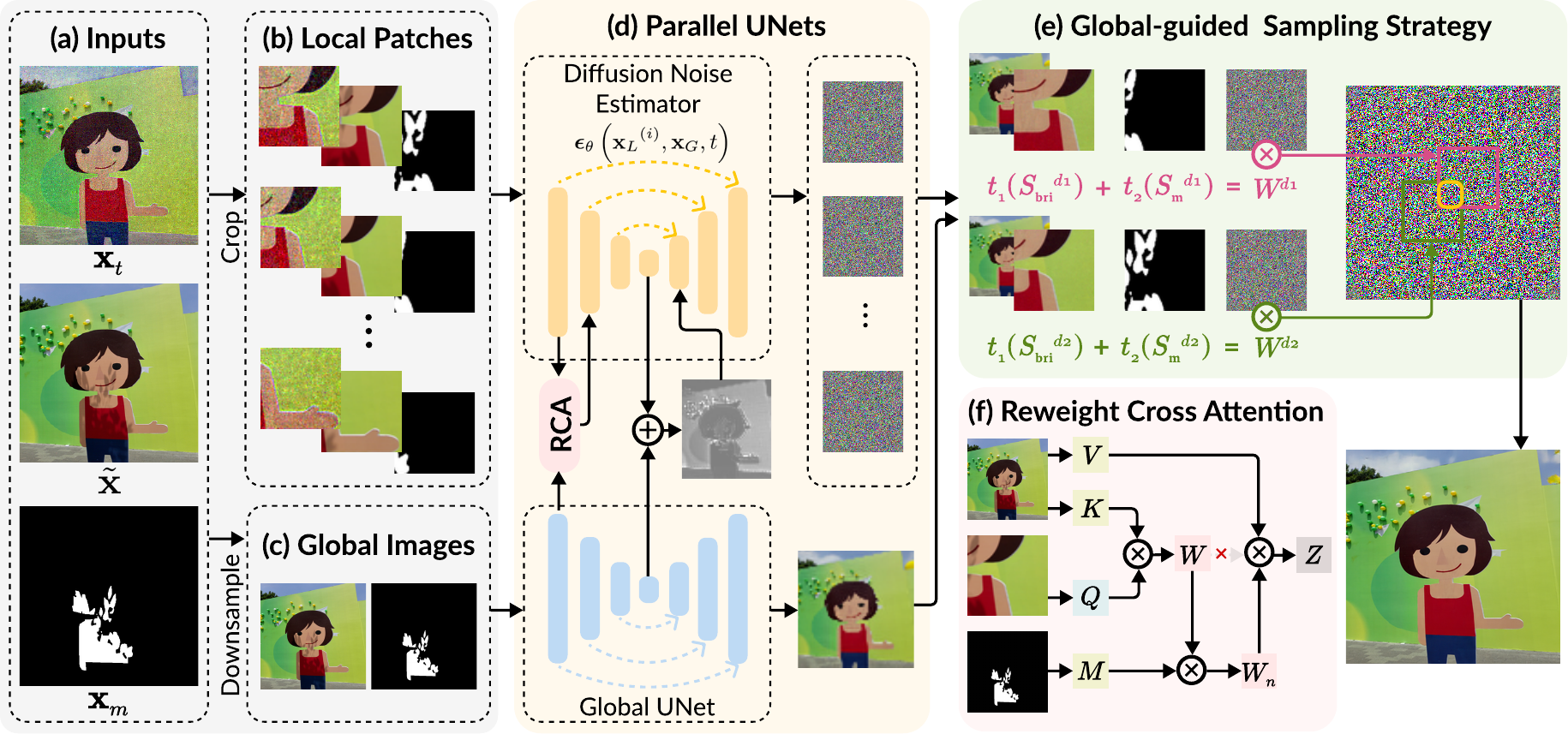}
    \caption{An overview of the Diff-Shadow. The local branch of Parallel UNets performs patch diffusion noise estimation using patches of the intermediate variable $\mathbf{x}_t$, the shadow image $\tilde{\mathbf{x}}$, and the shadow mask $\mathbf{x}_m$. The latter two images are also down-sampled as inputs of the global UNet, which constructs the low-resolution shadow-free image on the one hand and provides the global contextual information of non-shaded regions $\mathbf{x}_G$ into the local branch ${\mathbf{x}_L}^{(i)}$ using the proposed Reweight Cross Attention (RCA) module on the other hand. After estimating the noises of local patches, the Global-guided Sampling Strategy (GSS) works to construct the noise distribution for the whole image. $t_1$ and $t_2$ represent functions corresponding to the step $t$.}
    \label{fig:pipeline}
\end{figure*}

\noindent\textbf{Shadow and Non-Shadow Consistency.}
Shadow removal presents a unique challenge in that shadows corrupt only parts of an image, while abundant uncorrupted information remains in non-shadowed regions. However, previous methods~\cite{le2019shadow,le2020shadow} focused only on local shadow or non-shadow regions, leading to severe artifacts around shadow boundaries and inconsistent brightness between shadow and non-shadow areas. More recent approaches 
leverage contextual information from both shadow and non-shadow areas of the image to inform the shadow removal process. For instance, DeshadowNet~\cite{qu2017deshadownet} incorporates multi-scale information to expand the receptive field, leveraging texture and appearance information across the image. Chen~\emph{et al.} proposed CANet~\cite{chen2021canet}, using an external patch-matching module to discover texture relationships. Jin~\emph{et al.} proposed SG-ShadowNet~\cite{wan2022style}, reformulating shadow removal as an intra-image style transfer problem to harmonize de-shadowed regions with non-shadow areas.
Guo~\emph{et al.} introduced ShadowFormer~\cite{guo2023shadowformer}, incorporating a Shadow-Interaction Module in the transformer to model the global texture.
Related but divergent, we propose the Reweight Cross Attention module injected into the diffusion model,
empowering it with global textural relationship modeling.

\section{Method}
\label{sec:methods}

Given the intermediate variable $\mathbf{x}_\text{t}\in\mathbb{R}^{3\times H\times W}$, the shadow image $\tilde{\mathbf{x}}\in\mathbb{R}^{3\times H\times W}$ with corresponding shadow mask $\mathbf{x}_m\in\mathbb{R}^{1\times H\times W}$ in Fig.~\ref{fig:pipeline} (a), the proposed architecture is trained and inference using both high-resolution image patches as well as low-resolution down-sampled full images, which are optimized by the parallel UNets (Fig.~\ref{fig:pipeline} (d)) whose local branch performs the diffusion noise estimation of local patches and the global UNet recovers the down-sampled full shadow-free image.
The global information of non-shadow regions is integrated into the local branch through the RCA module  (Fig.~\ref{fig:pipeline} (f)) and the Convolutional Aggregation (CA) during training. The Global-guided Sampling Strategy (GSS) (Fig.~\ref{fig:pipeline} (e)) is utilized to combine the features of neighboring patches during inference. 

\subsection{Overall Architecture}

Condition diffusion model~\cite{saharia2022image} have achieved impressive performance for image-conditional data synthesis and editing tasks, which learns a conditional reverse process $p_\theta\left(\mathbf{x}_{0: T} \mid \tilde{\mathbf{x}}\right)$ without modifying the diffusion process $q\left(\mathbf{x}_{1: T} \mid \mathbf{x}_0\right)$ for ${\mathbf{x}}$, such that the sampled image has high fidelity to the data distribution conditioned on $\tilde{\mathbf{x}}$.
Common diffusive models train samples $\left(\mathbf{x}_0, \tilde{\mathbf{x}}\right) \sim q\left(\mathbf{x}_0, \tilde{\mathbf{x}}\right)$ from a paired data distributions (~\emph{e.g.}, a clean image $\mathbf{x}_0$ and a degraded image $\tilde{\mathbf{x}}$), and learn a conditional diffusion model that provides $\tilde{\mathbf{x}}$ to the reverse process:
\begin{equation}
\small
p_\theta\left(\mathbf{x}_{0: T} \mid \tilde{\mathbf{x}}\right)=p\left(\mathbf{x}_T\right) \prod_{t=1}^T p_\theta\left(\mathbf{x}_{t-1} \mid \mathbf{x}_t, \tilde{\mathbf{x}}\right).
\label{eq:conditional}
\end{equation}

As for shadow removal tasks with shadow masks $\mathbf{x}_m$, we design the parallel UNets architecture. The local branch performs the diffusion noise estimation and is conditioned by image patches: ${\mathbf{x}_L}^{(i)}=\left\{\tilde{{\mathbf{x}}}^{(i)},  {\mathbf{x}_m}^{(i)}\right\}$, where $\tilde{\mathbf{x}}^{(i)}=\operatorname{Crop}\left(\boldsymbol{L}_i \circ \tilde{\mathbf{x}}\right)$ and ${\mathbf{x}_m}^{(i)}=\text{Crop}\left(\boldsymbol{L}_i \circ \mathbf{x}_m\right)$ represent $R \times R$ patches, and $\text{Crop}(\cdot)$ extracts the patches from the location $\boldsymbol{L}_i$.  
The estimated noise of local patches will be combined during the inference through the proposed GSS module  (Fig.~\ref{fig:pipeline} (e)).
The global branch is optimized by the low-resolution down-sampled full images: $\mathbf{x}_G=\left\{\tilde{{\mathbf{x}}}',  {\mathbf{x}_m}'\right\}$. The contextual information of non-shadow regions plays an auxiliary role and is integrated into the local branch through the RCA module (Fig.~\ref{fig:pipeline} (f)).
Therefore, the reverse process in Eq.~\ref{eq:conditional} can be expanded as:
\begin{equation}
\footnotesize
p_\theta \! \left(\mathbf{x}_{0: T}^{(i)} \! \mid \! {\mathbf{x}_L}^{(i)}, \mathbf{x}_G\right) \! = \! p\left(\mathbf{x}_T^{(i)}\right) \prod_{t=1}^T p_\theta \! \left(\mathbf{x}_{t-1}^{(i)} \! \mid \! \mathbf{x}_t^{(i)}, {\mathbf{x}_L}^{(i)}, \mathbf{x}_G\right),
\end{equation}
where ${\mathbf{x}_0}^{(i)}=\text{Crop}\left(\boldsymbol{L}_i \circ \mathbf{x}_0\right)$. The training approach is outlined in Algorithm~\ref{alg:training}.

We can sample the intermediate $\mathbf{x}_t$ from the clean image $\mathbf{x}_0$ through $\mathbf{x}_t=\sqrt{\bar{\alpha}_t} \mathbf{x}_0+\sqrt{1-\bar{\alpha}_t} \boldsymbol{\epsilon}$, where $\alpha_t=1-\beta_t$, $\beta_t$ is a predefined variance schedule, $\bar{\alpha}_t=\prod_{i=1}^t \alpha_i$, and $\boldsymbol{\epsilon} \sim \mathcal{N}(\mathbf{0}, \mathbf{I})$.
The model is trained to predict the noise by:
\begin{equation}
\small
\boldsymbol{{\epsilon}_t}=\boldsymbol{\epsilon}_\theta\left(\sqrt{\bar{\alpha}_t} {\mathbf{x}_0}^{(i)}+\sqrt{1-\bar{\alpha}_t} \boldsymbol{\epsilon}, {\mathbf{x}_L}^{(i)}, \mathbf{x}_G, t\right).
\end{equation}

\begin{algorithm}[t]
\caption{Diffusive shadow removal model training}
\label{alg:training}
\begin{algorithmic}[1]
\small
\renewcommand{\algorithmicrequire}{\textbf{Input:}}
\renewcommand{\algorithmicensure}{\textbf{Output:}}
\REQUIRE shadow image $\tilde{\mathbf{x}}$, clean image $\mathbf{x}_0$ and mask $\mathbf{x}_m$
\REPEAT
\STATE Randomly sample a patch $\boldsymbol{L}_i$
\STATE { $\mathbf{x}_0^{(i)}=\operatorname{Crop}\left(\boldsymbol{L}_i \circ \mathbf{x}_0\right)$ }
\STATE 
$\mathbf{x}_L{ }^{(i)}=\left\{\tilde{\mathbf{x}}^{(i)}, \mathbf{x}_m{ }^{(i)}\right\}$ where $\tilde{\mathbf{x}}^{(i)}=\operatorname{Crop}\left(\boldsymbol{L}_i \circ \tilde{\mathbf{x}}\right)$ and $\mathbf{x}_m^{(i)}=\operatorname{Crop}\left(\boldsymbol{L}_i \circ \mathbf{x}_m \right)$
\STATE 
$\mathbf{x}_G = \left\{\tilde{\mathbf{x}}', {\mathbf{x}_m}'\right\}$ where $\tilde{\mathbf{x}}' = \operatorname{Dowmsample}(\tilde{\mathbf{x}})$ and ${\mathbf{x}_m}' = \operatorname{Dowmsample}(\mathbf{x}_m)$
\STATE $t \sim \text{Uniform}\{1,\ldots,T\}$
\STATE $\boldsymbol{\epsilon} \sim \mathcal{N}(\mathbf{0}, \mathbf{I})$
\STATE Perform a gradient descent step for \\ \qquad 
{\footnotesize
$\nabla_\theta\left\|\boldsymbol{\epsilon}_t-\boldsymbol{\epsilon}_\theta\left(\sqrt{\bar{\alpha}_t} \mathbf{x}_0^{(i)}+\sqrt{1-\bar{\alpha}_t} \boldsymbol{\epsilon}, {\mathbf{x}_L}^{(i)}, \mathbf{x}_G, t\right)\right\|^2$
}
\UNTIL converged
\RETURN $\theta$
\end{algorithmic} 
\end{algorithm}

\subsection{Reweight Cross Attention}\label{sec:RCA}

Existing patch-based methods lack generality since the regions inside a patch may all be corrupted. Apart from the local patch-based diffusion branch, we introduce a parallel global UNet to recover low-resolution shadow-free images.
The Reweight Cross Attention (RCA) module is introduced to integrate the global information into the local patch by combining the low-level features. As shown in Fig.~\ref{fig:pipeline} (f), the output
of RCA is:
\begin{equation}
\mathbf{Z}=\operatorname{softmax}\left(\frac{\mathbf{Q} \mathbf{K}^T}{d}\right)\mathbf{M}\mathbf{V},
\end{equation}
 $\mathbf{Q}$ represents the projected queries of patch features, $\mathbf{K}$ and $\mathbf{V}$ are the projected keys and values of global image features, $d$ is the scaling parameter, $\mathbf{M}$ is the binary mask to multiply with the features, and $\mathbf{Z}$ represents the output interacted features. Attention maps between $\mathbf{Q}$ and $\mathbf{K}$ are first calculated to capture the mutual information between pixels in the local branch and regions in the global branch. Subsequently, the mask is introduced to assist in suppressing mutual information values in the global shadow regions. The adjusted attention map is then multiplied by the global $\mathbf{V}$ vector to obtain the expected global information for the local branch.

We interact the low-level features of the first convolutional block in the global UNet with the corresponding layer in the local UNet through the RCA because low-level features concentrate more on image edges, colors, and corners, which are important for shadow removal tasks. As for high-level abstract features, we apply a simple concatenation to achieve interaction. In contrast to recent cross-attention modules in ShadowFormer \cite{guo2023shadowformer} whose $\mathbf{Q}$, $\mathbf{K}$ and $\mathbf{V}$ originate from the same global feature, the  RCA integrates both local and global information. Furthermore, other methods boost the mutual information between shadow and non-shadow pairs through masks. However, the shadow information is dispensable for recovering non-shadow regions. Contrarily, our attention mechanism utilizes masks to suppress the global shadow values through the attention map, integrating the more important non-shadow information from the global branch into the local branch.

\begin{figure}[t]
    \centering
    \includegraphics[width=1.0\linewidth]{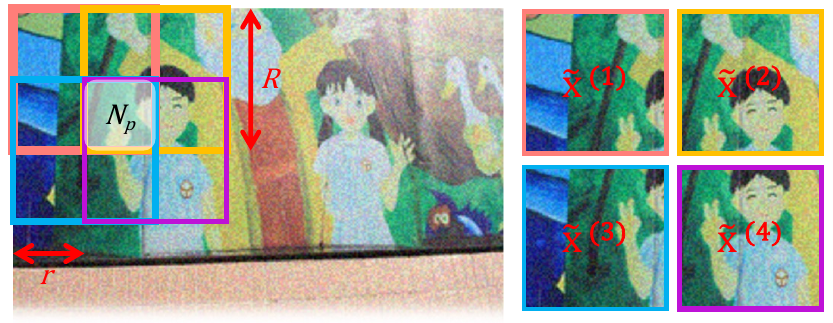}
    \caption{The illustration of noise merge for overlapping patches. $N_p$ represents the number of overlapping patches for each pixel. $r$ and $R$ are the step size and the patch size.
    }
    \label{fig:gnm}
\end{figure}

\subsection{Global-guided Samping Strategy}\label{sec:noise_merge}

To overcome the shortcomings of merging artifacts from independently intermediate patches, Ozdenizci~\emph{et al.} proposed to perform reverse sampling based on the mean estimated noise for each pixel in overlapping patch regions, which works better for weather image with relatively unified degradation~\cite{ozdenizci2023restoring}. 
However, as for shadow removal tasks with extreme inconsistencies between shadow and non-shadow regions, predicting mean estimated noise lacks ability to achieve a consistent illumination distribution. Therefore, we propose the Global-guided Sampling Strategy (GSS) to maintain 
illumination consistency. 

The general idea of patch-based restoration is to operate locally on patches extracted from the image and optimally merge the results. 
We consider a grid-like arranged parsing scheme over the intermediate variable $\mathbf{x}_t$, the shadow image $\tilde{\mathbf{x}}$ and the shadow mask $\mathbf{x}_m$, and extract all patches by moving over the $R \times R$ grid with a step size of $r$ in both horizontal and vertical dimensions, as shown in Fig.~\ref{fig:gnm}. During patch-based diffusive image restoration,
we perform noise estimation based on adaptive estimated noise for each pixel in overlapping patch regions, at any given denoising time step $t$. As shown in Fig.~\ref{fig:pipeline} (e), the weight of each patch is determined by two parts: the brightness-based score $S_\text{bri}$ and the mask-based score $S_\text{m}$. The former one considers the brightness consistency of the local patches and the global images, while the latter one corresponds to the proportion of shadow and non-shadow regions. 

The brightness-based score $S_\text{bri}$ constrains that the patch recovered by the local branch should show similar brightness with the full image generated by the global branch.
The $S_\text{bri}$ is defined as:
\begin{equation}
\small
S_\text{bri}=1-\frac{\mathrm{avg}{(|L_\text{gray}-G_\text{gray}|)}}{\mathrm{max}({(|L_\text{gray}-G_\text{gray}|)})},
\end{equation}
where $L_\text{gray}$ is the recovered gray image of the local patch, $G_\text{gray}$ is the recovered gray image of global image. 

In addition, the masked-based score $S_\text{m}$ pays more attention to non-shadow regions. When the non-shadow region dominates the image patch, the receptive field of the local branch encompasses both shadow and non-shadow regions, providing a valuable way to eliminate shadows through non-shadow information. Conversely, when the non-shadow region dominates the image patch, the predicted illumination may mainly originate from the global branch, whose performance tends to be unsatisfactory. 
Based on the binary mask, the ratio of non-shadow region among the patch can be calculated through: $R_{\text{sf}}=P_\text{sf}/P_\text{total}$, where $P_\text{sf}$ is the number of non-shadow pixels and $P_\text{total}$ is the number of total patch pixels. The mask-based score $S_\text{m}$ is defined as:
\begin{equation}
\small
S_\text{m}=\left\{\begin{array}{cc}
1 & R_{\text{sf}}\ge T_{\text{sf}}\\
R_{\text{sf}}/{T_{\text{sf}}} & R_{\text{sf}}<T_{\text{sf}}
\end{array},\right.
\end{equation}
where $T_{\text{sf}}$ is the threshold of the ratio. In our implementation, $T_{\text{sf}}$ is set to 0.5 and $\mathrm{eps}$ is set to 0.01.

The weights of each patch contributed to constructing the full noise map are calculated as:
\begin{equation}
\small
W^{d}=\frac{t \cdot S_\text{bri}}{T} + \frac{(T-t) \cdot S_\text{m}}{T},
\end{equation}
where $T$ is the total steps. The final adaptive estimated noise is computed as:
\begin{equation}
\small
\frac{1}{\sum_{i}  W^{d_i}} \sum_{i=1}^{N_p}\left(W^{d_i} \! \cdot \!\epsilon_\theta\left(\mathbf{x}_t^{(i)}, {\mathbf{x}_L}^{(i)}, {\mathbf{x}_G}, t\right)\right),
\end{equation}
where $i \in  \left \{ 1, \ldots, N_p  \right \}$ and $N_p$ is the number of overlapping patches of each pixel. The test time diffusive image restoration
method is outlined in Algorithm~\ref{alg:inference}.

\begin{algorithm}[t]
\caption{Global-guided diffusive image restoration}
\label{alg:inference}
\small
\begin{algorithmic}[1]
\renewcommand{\algorithmicrequire}{\textbf{Input:}}
\renewcommand{\algorithmicensure}{\textbf{Output:}}
\REQUIRE shadow image $\tilde{\mathbf{x}}$, shadow mask $\mathbf{x}_m$, conditional diffusion model $\boldsymbol{\epsilon}_{\theta}$, number of implicit sampling steps $S$, dictionary of $D$ overlapping patch locations.
\STATE $\mathbf{x}_t \sim \mathcal{N}(\mathbf{0}, \mathbf{I})$
\FOR {$i = S,\ldots,1$}
\STATE $t = (i-1)\cdot T / S + 1$
\STATE $t_{\text {next }}=(i-2) \cdot T / S+1$ if $i>1$ else 0
\STATE $\hat{\boldsymbol{\Omega}}_t = 0$ and $M = 0$
\FOR {$d = 1,\ldots,D$}
\STATE $\mathbf{x}_t^{(d)}=\operatorname{Crop}\left(\boldsymbol{L}_d \circ \mathbf{x}_t\right)$
\STATE 
$\mathbf{x}_L{ }^{(d)}\!=\!\left\{\tilde{\mathbf{x}}^{(d)}, \mathbf{x}_m{ }^{(d)}\right\}$ where
$\tilde{\mathbf{x}}^{(d)}\!=\!\operatorname{Crop}\left(\boldsymbol{L}_d \circ \tilde{\mathbf{x}}\right)$ and ${\mathbf{x}_m}^{(d)}\!=\!\operatorname{Crop}\left(\boldsymbol{L}_d \circ \mathbf{x}_m\right)$
\STATE 
$\mathbf{x}_G = \left\{\tilde{\mathbf{x}}', {\mathbf{x}_m}'\right\}$ where 
$\mathbf{x}' = \operatorname{Dowmsample}(\mathbf{x})$ and ${\mathbf{x}_m}' = \operatorname{Dowmsample}(\mathbf{x}_m)$
\STATE $\hat{\boldsymbol{\Omega}}_t=\hat{\boldsymbol{\Omega}}_t+ W^{d} \cdot \boldsymbol{\epsilon}_\theta\left(\mathbf{x}_t^{(d)}, \tilde{\mathbf{x}_L}^{(d)}, \mathbf{x}_G, t\right)$\\
\STATE $M = M + W^{d}$\\
\ENDFOR
\STATE $\hat{\boldsymbol{\Omega}}_t=\hat{\boldsymbol{\Omega}}_t / M $ \\
\STATE $\mathbf{x}_t \leftarrow \sqrt{\bar{\alpha}_{t_{\text {next }}}}\left(\frac{\mathbf{X}_t-\sqrt{1-\bar{\alpha}_t} \cdot \hat{\boldsymbol{\Omega}}_t}{\sqrt{\bar{\alpha}_t}}\right)+\sqrt{1-\bar{\alpha}_{t_{\text {next }}}} \cdot \hat{\boldsymbol{\Omega}}_t$
\ENDFOR
\RETURN $\mathbf{x}_t$
\end{algorithmic} 
\end{algorithm}

\subsection{Optimization}

The Diff-Shadow involves two parallel UNet networks operating the local patch optimization and the global shadow-free image reconstruction. The objective function includes the following two items: the diffusive objective function $L_\text{diff}$ and the global loss function $L_\text{global}$. 
Total loss $\mathcal{L}_{\text {total }}=\mathcal{L}_{\text {diff }}+\lambda \mathcal{L}_{\text {global}}$, where $\lambda$ is a hyper-parameter to balance the contributions of the two losses. We empirically set it to 1.

\noindent\textbf{The Diffusive Objective Function.} Refer to~\cite{ho2020denoising}, the $\mathcal{L}_\text{diff}$ is defined as:
\begin{equation}
\small
\mathcal{L}_\text{diff}=\mathbb{E}_{\mathbf{x}_0, t, \boldsymbol{\epsilon}}\left\|\boldsymbol{{\epsilon}}_t-\boldsymbol{\epsilon}\right\|_F^2.
\end{equation}.

\noindent\textbf{The Global Loss Function.} The global UNet recovers shadow-free images from down-sampled shadow images and mask images:
\begin{equation}
\small
\mathcal{L}_{\text {global}}=\mathbb{E}_{\mathbf{{x}_0}', \mathbf{{x}_m}'}\left\|\mathbf{x}_r-\mathbf{{x}_0}'\right\|_F^2,
\end{equation}
where $\mathbf{x}_r$ is the global output, $\mathbf{{x}_0}'$ and $\mathbf{{x}_m}'$ are the down-sampled clean images and shadow mask, respectively.

\section{Experiments}
\label{sec:experiments}

\subsection{Experimental Setups}
\noindent\textbf{Datasets.}
We used three standard datasets: 1) ISTD dataset~\cite{wang2018stacked} includes 1,330 training and 540 testing triplets (shadow images, masks and shadow-free images); 2) Adjusted ISTD (ISTD+) dataset~\cite{le2019shadow} reduces the illumination inconsistency between the shadow and shadow-free image of ISTD; 3) SRD dataset~\cite{qu2017deshadownet} consists of 2,680 training and 408 testing pairs of shadow and shadow-free images without the shadow masks. Referring to ShadowDiffusion, we applied the prediction mask provided by DHAN~\cite{cun2020towards} for the training and testing.

\noindent\textbf{Implementation Details.}
Our Diff-Shadow is trained using eight NVIDIA GTX 2080Ti GPUs. The Adam optimizer~\cite{kingma2014adam} is applied to optimize the parallel UNets with a fixed learning rate of $lr=2e^{-4}$ without weight decay and the training epoch is set as 2,000. The exponential moving average~\cite{nichol2021improved} with a weight of 0.999 is applied for parameter updating. Further details regarding the inference time and network complexity can be found in the supplementary.

\noindent\textbf{Training Specifications.}
For each dataset, the images are resized into 256 $\times$ 256 for training. For each iteration, we initially sampled 8 images from the training set and randomly cropped 16 patches with size 64 $\times$ 64 from each image plus a down-sampled global image with the same size corresponding to each patch. This process resulted in mini-batches consisting of 96 samples. We used input time step embeddings for $t$ through sinusoidal positional encoding~\cite{vaswani2017attention} and provided these embeddings as inputs to each residual block in the local and global branches, allowing the model to share parameters across time. 
During sampling, the patch size $R$ is set to 64 and the step size $r$ is set to 8 for covering the whole image.

\noindent\textbf{Evaluation Measures.}
Following previous works~\cite{fu2021auto, jin2021dc, zhu2022bijective, guo2023shadowformer, guo2023shadowdiffusion}, we utilized the Root Mean Square Error (RMSE) in the LAB color space as the quantitative evaluation metric of the shadow removal results, comparing to the ground truth shadow-free images. Lower RMSE scores correspond to better performance. In addition, we also adopt the common Peak Signal-to-Noise Ratio (PSNR) and the Structural Similarity Index Measure (SSIM)~\cite{wang2004image} to measure the performance in the RGB color space. 
\begin{table*}[ht!]
\centering
\resizebox{1.0\linewidth}{!}{
\begin{tabular}{c|c|ccc|ccc|ccc}
\toprule
\multirow{2}{*}{Dataset} & \multirow{2}{*}{Method}                                                            & \multicolumn{3}{c|}{Shadow Regions (S)}                                                                          & \multicolumn{3}{c|}{Non-Shadow Regions (NS)}                                                                    & \multicolumn{3}{c}{All Image (All)}                                                                            \\ 
                         &  & PSNR$\uparrow$ & SSIM$\uparrow$ & RMSE$\downarrow$ & PSNR$\uparrow$ & SSIM$\uparrow$ & RMSE$\downarrow$ & PSNR$\uparrow$ & SSIM$\uparrow$ & RMSE$\downarrow$ \\
\midrule
\multirow{10}{*}{ISTD}                  & Input Image                                                                        & 22.40                              & 0.936                              & 32.10                                 & 27.32                              & 0.976                              & 7.09                                  & 20.56                              & 0.893                              & 10.88                                 \\
                                        & Guo~\emph{et al.} & 27.76                              & 0.964                              & 18.65                                 & 26.44                              & 0.975                              & 7.76                                  & 23.08                              & 0.919                              & 9.26                                  \\
                                        & ST-CGAN                               & 33.74                              & 0.981                              & 9.99                                  & 29.51                              & 0.958                              & 6.05                                  & 27.44                              & 0.929                              & 6.65                                  \\
                                        & DHAN                                   & 35.53                              & 0.988                              & 7.49                                  & 31.05                              & 0.971                              & 5.30                                  & 29.11                              & 0.954                              & 5.66                                  \\
                                        & Fu~\emph{et al.}     & 34.71                              & 0.975                              & 7.91                                  & 28.61                              & 0.880                              & 5.51                                  & 27.19                              & 0.945                              & 5.88                                  \\
                                        & DC-ShadowNet                                & 31.69                              & 0.976                              & 11.43                                 & 28.99                              & 0.958                              & 5.81                                  & 26.38                              & 0.922                              & 6.57                                  \\
                                        & Zhu et al.                                & 36.95                              & 0.987                              & 8.29                                 & 31.54                              & 0.978                              & 4.55                                  & 29.85                              & 0.960                              & 5.09                                  \\
                                        & ShadowFormer                      & 38.19                              & \underline{0.991}                              & 5.96                                  & \underline{34.32}                              & \underline{0.981}                              & 3.72                                  & 32.21                              &  0.968                              & 4.09                                  \\
                                        & ShadowDiffusion                & \textbf{40.15}    & \textbf{0.994}    & \underline{4.13}                                  & 33.70                             & 0.977                              & \underline{4.14}                                  & \underline{32.33}                            & \underline{0.969}                              & \underline{4.12}                                  \\
                                        & \textbf{Ours}                                                     & \underline{39.71}                              & \textbf{0.994}                              & \textbf{1.41}        & \textbf{35.58}    & \textbf{0.987}    & \textbf{2.48}       & \textbf{33.69}    & \textbf{0.976}    & \textbf{2.89}        \\
\midrule
\multirow{10}{*}{SRD}                   & Input Image                    & 18.96                              & 0.871                              & 36.69                                 & 31.47                              & 0.975                              & 4.83                                  & 18.19                              & 0.830                              & 14.05                                 \\
                                        & Guo~\emph{et al.} & -                                  & -                                  & 29.89                                 & -                                  & -                                  & 6.47                                  & -                                  & -                                  & 12.60                                 \\
                                        & DeshadowNet                        & -                                  & -                                  & 11.78                                 & -                                  & -                                  & 4.84                                  & -                                  & -                                  & 6.64                                  \\
                                        & DHAN                                   & 33.67                              & 0.978                              & 8.94                                  & 34.79                              & 0.979                              & 4.80                                  & 30.51                              & 0.949                              & 5.67                                  \\
                                        & Fu~\emph{et al.}     & 32.26                              & 0.966                              & 9.55                                  & 31.87                              & 0.945                              & 5.74                                  & 28.40                              & 0.893                              & 6.50                                  \\
                                        & DC-ShadowNet                                & 34.00                              & 0.975                              & 7.70                                  & 35.53                              & 0.981                              & 3.65                                  & 31.53                              & 0.955                              & 4.65                                  \\
                                        & BMNet                                & 35.05                              & 0.981                              & 6.61                                  & 36.02                              & 0.982                              & 3.61                                  & 31.69                              & 0.956                              & 4.46                                  \\
                                        & SG-ShadowNet                         & -                                  & -                                  & 7.53                                 & -                                  & -                                  & 2.97                                 & -                                  & -                                  & 4.23                                  \\
                                        & Zhu et al.                                & 34.94                              & 0.980                             & 7.44                                 & 35.85                              & 0.982                              & 3.74                                  & 31.72                              & 0.952                             & 4.79                                  \\
                                        & ShadowFormer                      & 36.91                              & \textbf{0.989}    & 5.90                                  & 36.22                              & \underline{0.989}                              & 3.44                                  & 32.90                              & \underline{0.958}                              & 4.04                                  \\
                                        & ShadowDiffusion                & \textbf{38.47}                              & 0.987                              & \underline{4.98}                                  & \underline{37.78}                              & 0.985                              & \underline{3.44}                                  & \underline{34.73}                              & 0.957                              & \underline{3.63}                                  \\
                                        & \textbf{Ours}                                                     & \underline{37.91}         & \underline{0.988}         & \textbf{1.81}            & \textbf{39.49}         & \textbf{0.990}         & \textbf{1.65}            & \textbf{34.93}         & \textbf{0.977}         & \textbf{2.52}            \\
\midrule
\multirow{9}{*}{ISTD+}   & Input Image                                                                    & 18.96                               & 0.871                               & 36.69                                 & 31.47                               & 0.975                               & 4.83                                  & 18.19                               & 0.830                               & 14.05                                 \\
                         & ST-CGAN                           & -                                   & -                                   & 13.40                                  & -                                   & -                                   & 7.70                                   & -                                   & -                                   & 8.70                                  \\
                         & DeshadowNet                     & -                                   & -                                   & 15.90                                  & -                                   & -                                   & 6.00                                   & -                                   & -                                   & 7.60                                  \\
                         & Fu~\emph{et al.} & 36.04                               & 0.976                               & 6.60                                   & 31.16                               & 0.876                               & 3.83                                   & 29.45                               & 0.840                               & 4.20                                   \\
                         & DC-ShadowNet                            & 32.20                               & 0.976                               & 10.43                                 & 33.21                               & 0.963                               & 3.67                                  & 28.76                               & 0.922                               & 4.78                                  \\
                         & BMNet                            & 37.30                               & \underline{0.990}                                   & 6.19                                   & 35.06                               & \underline{0.974}                                   & 3.09                                   & 32.30                               & 0.955                               & 3.60                                   \\
                         & SG-ShadowNet                        & -                                  & -                                  & 5.9                                 & -                                         & -                                  & 2.90                                 & -                                  & -                                  & 3.4     \\
                         & ShadowFormer                  & 39.67                               & -                                   & 5.20                                   & 38.82                               & -                                   & 2.30                                   & 35.46                               & -                                   & 2.80                                   \\
                         & ShadowDiffusion            & \underline{39.82}                               & -                                   & \underline{4.90}                                   & \underline{38.90}                               & -                                   & \underline{2.30}                                   & \underline{35.68}                               & \underline{0.970}                               & \underline{2.70}                                   \\
                         & \textbf{Ours}                                                 & \textbf{40.63}     & \textbf{0.993}     & \textbf{1.30}        & \textbf{39.69}    & \textbf{0.989}     & \textbf{1.57}        & \textbf{36.47}     & \textbf{0.979}     & \textbf{2.09}        \\ 
\bottomrule
\end{tabular}
}
\caption{The quantitative comparisons on ISTD, SRD, and ISTD+ datasets. `-' represents the values that were not reported.}
\label{table:1}
\end{table*}
\begin{figure*}[!ht]
    \centering
    \includegraphics[width=0.95\linewidth]{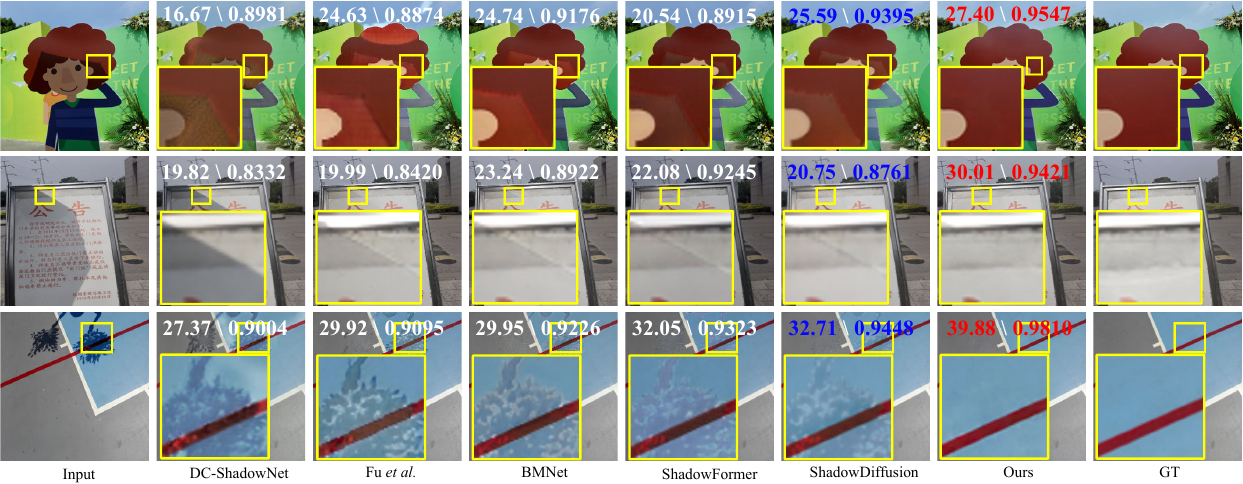} 
    \caption{Qualitative comparisons on ISTD+.
    The values 
    represent the `PSNR/SSIM'.}
    \label{fig:comp}
\end{figure*}

\subsection{Comparison with the State-of-the-art}
We compare the proposed Diff-Shadow with the popular or state-of-the-art shadow removal algorithms, including one traditional method: Guo~\emph{et al.}~\cite{guo2012paired}, several deep learning-based methods: DeshadowNet~\cite{qu2017deshadownet}, ST-CGAN~\cite{wang2018stacked}, DHAN~\cite{cun2020towards}, Fu~\emph{et al.}~\cite{fu2021auto}, DC-ShadowNet~\cite{jin2021dc}, BMNet~\cite{zhu2022bijective}, ShadowFormer~\cite{guo2023shadowformer}, and one diffusion-based method: ShadowDiffusion~\cite{guo2023shadowdiffusion}. The shadow removal results of the competing methods are quoted from the original papers. Following the previous methods~\cite{guo2023shadowformer, guo2023shadowdiffusion},
we evaluate the removal results with 256 $\times$ 256 resolution, while the results for other resolutions are detailed in the supplementary.

\noindent\textbf{Quantitative Evaluation.}
Table~\ref{table:1} shows the quantitative results on the test sets over ISTD, SRD, and ISTD+, respectively. The superiority of the proposed method over competing approaches is evident across all three datasets, as it consistently achieves superior performance. 
Thanks to the powerful modeling capabilities of the diffusion model for images, the proposed method improves the PSNR of the SRD from 32.90 dB to 34.93 dB when compared with the transformer-based method ShadowFormer. Furthermore, compared to ShadowDiffusion which is also a diffusion-based method, our parallel network architecture interacts the local and global information more effectively. The novel Global-guided Sampling Strategy (GSS) further enhances the accuracy of shadow-free image restoration, which results in an improvement in PSNR from 32.33 dB to 33.69 dB on the ISTD.
In summary, our Diff-Shadow achieves new state-of-the-art results on three widely-used benchmarks.

\noindent\textbf{Qualitative Evaluation.}
We present the qualitative comparison in Fig.~\ref{fig:comp} and the supplementary. Diff-Shadow stands out among previous methods for its ability to remove shadows, while preserving the surrounding image textures and luminance, even in areas of high contrast that pose challenges for shadow removal, ~\emph{e.g.}, in the third example of Fig.~\ref{fig:comp}, there is a distinct color disparity between the ground and the shadow. Previous methods struggled to remove the fragmented shadows of branches at this location, resulting in uneven color distribution or incomplete removal results. The result of our method is the most similar to the ground truth.
\begin{table}[ht!]
\centering
\begin{tabular}{l|ccc}
\toprule
Methods     & S       & NS       & ALL  \\
\midrule
(a) Single-head(w/o global)  &2.64         &3.41         & 3.36               \\
(b) Single-head(w global)  & 2.55        &  3.19        & 3.04                \\
(c) Ours (w/o RCA) &2.31       & 2.14      & 2.73                           \\
(d) Ours (complete model)    & \textbf{1.81}      & \textbf{1.65}     & \textbf{2.52} \\ 
\bottomrule
\end{tabular}
\caption{Ablation studies of different network structures on SRD dataset. We report RMSE values.}
\label{table:methods_ablation}
\end{table}

\begin{table}[hh!]
\centering
\begin{tabular}{l|ccc}
\toprule
Sampling    & S       & NS       & ALL \\
\midrule
(a) w/o  GSS    & 2.97        & 3.63        & 3.81          \\
(b) $W^d=1$   &2.64         &  3.42        & 3.27           \\
(c) $S_\text{bri}$   & 2.16              & 2.99               &  2.96                       \\
(d) $S_\text{bri}+S_\text{m}$     & \textbf{1.81}      & \textbf{1.65}     & \textbf{2.52}\\ 
\bottomrule
\end{tabular}
\caption{Ablation studies of different sampling strategies on SRD dataset. We report RMSE values.}
\label{table:Sampling_ablation}
\end{table}
\begin{figure}[h!]
    \centering
    \includegraphics[width=1.0\linewidth]{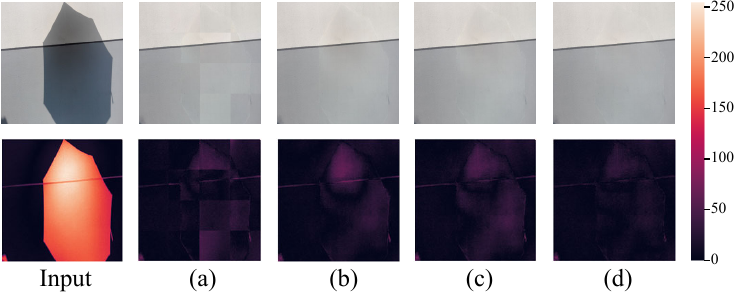}
    \caption{Visual examples of results and error maps for ablation study with different sampling strategies, along with corresponding results from Table~\ref{table:Sampling_ablation}.}
    \label{fig:ablation2}
\end{figure}
\subsection{Ablation Studies}

\noindent\textbf{Effectiveness of Parallel UNets.}
The global branch is 
removed to only accommodate the local inputs. Consequently, the network degrades to a single-head UNet structure. As in Table~\ref{table:methods_ablation} (a), the single-head UNet shows the worst performance due to missing global context information. We tried to concatenate the patch and global images together at the network input level, and the results in Table~\ref{table:methods_ablation} (b) show that the introduction of global context information works better than patch input alone.
Then, to verify the effectiveness of the RCA module, we retained the two-branch network mechanism but eliminated the RCA module from the first layer and kept only the concatenate connection in the middle layer. As shown in Table~\ref{table:methods_ablation} (c), the RCA module is effective in interacting two branches and enhancing the final result.

\noindent\textbf{Effectiveness of Global-guided Sampling Strategy.}
Table~\ref{table:Sampling_ablation} (a) presents the experiments without overlapping between patches.
It turns out significant boundaries between patches in Fig.~\ref{fig:ablation2} (a).
Second, we established the adaptive noise weights as fixed $W^d=1$ in Table~\ref{table:Sampling_ablation} (b), which leads to a degradation of the sampling procedure.
Reflected in Fig.~\ref{fig:ablation2} (b), we can see that there is a large discrepancy between the results and the ground truth.
Third, we controlled the noise merge weights solely by the difference between the recovered patch brightness and the recovered global image brightness,~\emph{i.e.}, only brightness-based score $S_\text{bri}$ exists. The qualitative patch image restored at the beginning of the sampling phase is heavily noisy in Fig.~\ref{fig:ablation2} (c). Therefore, it is not accurate to control the weights using brightness alone. The results in Table~\ref{table:Sampling_ablation} (c) are inferior to those of our full strategy in Table~\ref{table:Sampling_ablation} (d), which considers both the proportion of shaded areas in the patch and the brightness difference between the restored global and local image.

\section{Conclusion} \label{sec:conclusion}
In this work, we propose a global-guided diffusion model named Diff-Shadow, for high-quality shadow removal with no obvious boundaries and consistent illumination. We design a parallel network structure in the framework of the diffusion model to assist in the recovery of local patches using the shadow-free regions from the global image.
In addition, we propose a sampling strategy named GSS to further improve the performance of shadow removal during diffusion denoising. Comprehensive experiments demonstrate that Diff-shadow outperforms the state-of-the-art.

\appendix
\section{Acknowledgments}
This work was supported in part by the National Natural Science Foundation of China (NSFC) under Grant Nos. 62372091, 62071097, 62402136 and in part by Sichuan Science and Technology Program under Grant Nos. 2023NSFSC0462, 2023NSFSC0458, 2023NSFSC1972.

\bibliography{aaai25}

\clearpage
\section{Diff-Shadow: Global-guided Diffusion Model for Shadow Removal Supplementary}
In the supplementary material, we first provide more implementation details of the proposed Diff-Shadow.

Then, we present some visual results on complex scenes, and more visual comparisons on the ISTD+~\cite{le2019shadow} and SRD~\cite{qu2017deshadownet} datasets to demonstrate the effectiveness of our approach. 
Further, we provide the analysis of differences with current methods.
Finally, we discuss some limitations of the proposed method.

\section{Implementation Details}
\label{sec:details}
In our implementation, we employed a consistent network structure, as described in Table~\ref{table:Network}. Our parallel network consists of two UNet structures.
The local and global branches are nearly identical, with the only difference being the absence of a self-attention block in the global branch. The network incorporates six-scale resolutions and includes two residual blocks per-resolution. Detailed configurations and parameters can be found in Table~\ref{table:Network}.

In the local branch, we perform a channel-wise concatenation of the patch of the shadow image $\tilde{{\mathbf{x}}}^{(i)}$, the variable ${\mathbf{x}_t}^{(i)}$, and the mask ${\mathbf{x}_m}^{(i)}$, resulting in seven-dimensional input channels (RGB for $\tilde{{\mathbf{x}}}^{(i)}$ and ${\mathbf{x}_t}^{(i)}$, and gray channel for ${\mathbf{x}_m}^{(i)}$). Additionally, in the global branch, we concatenate the down-sampled full shadow image $\tilde{\mathbf{x}}'$ and the mask ${\mathbf{x}_m}'$, resulting in four-dimensional input image channels. Therefore, our network has a total of eleven input channels.

For the diffusion model, detailed configurations and parameters can be found in Table~\ref{table:Diffusion}.

\section{More Results}
\label{sec:complex}
\begin{figure}[ht!]
    \centering
    \includegraphics[width=1.0\linewidth]{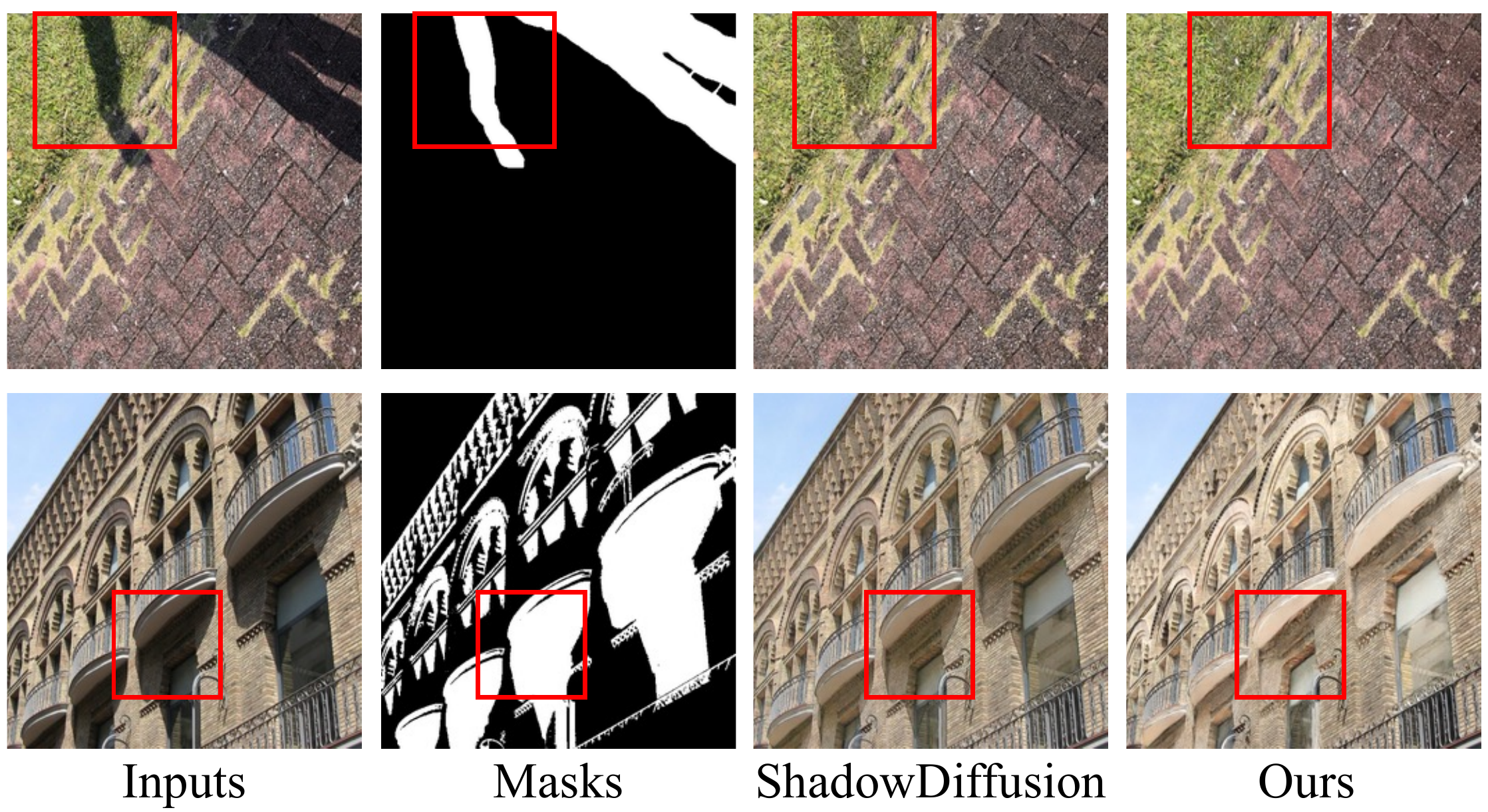} 
    \caption{Qualitative results on CUHK-Shadow dataset.}
    \label{fig:comp_cuhk}
\end{figure}

\begin{figure}[ht!]
    \centering
    \includegraphics[width=1.0\linewidth]{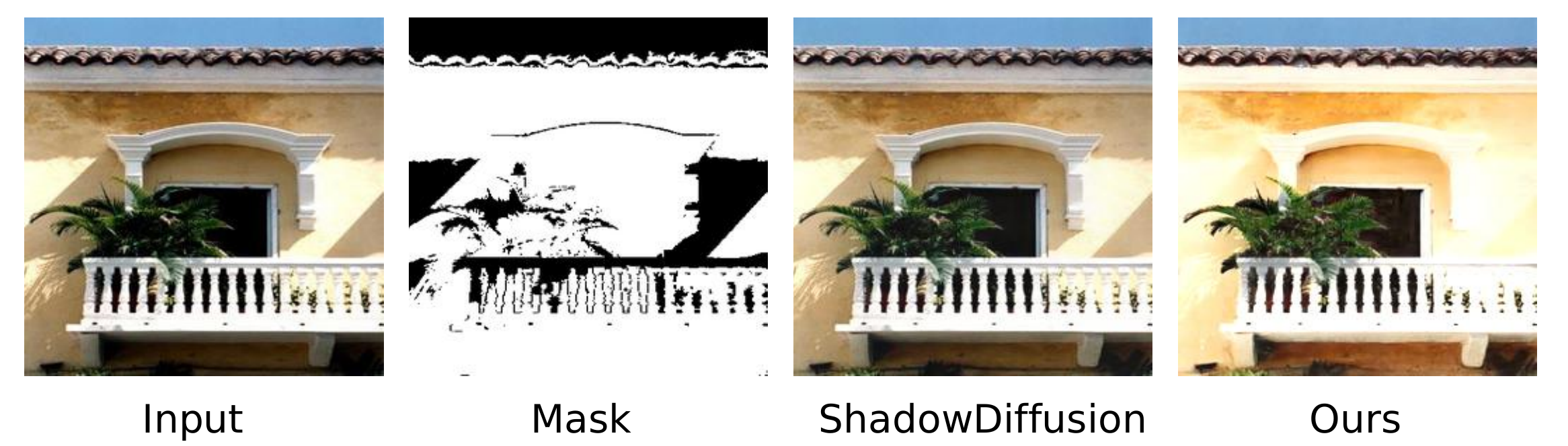} 
    \caption{Failure case on CUHK-Shadow dataset.}
    \label{fig:comp_failure}
\end{figure}

\noindent\textbf{Comparisons with ShadowDiffusion on Complex Dataset.}
Table~\ref{table:comp_with shadowdiffusion} presents the quantitative comparison with ShadowDiffusion on the original resolution. In addition, the complex CUHK-Shadow dataset focuses on the segmentation of shadows and does not contain shadow-free ground truth. Figure~\ref{fig:comp_cuhk} shows the visual results of generalizing the models trained on the SRD dataset to CUHK-Shadow dataset for testing. However, as shown in Figure~\ref{fig:comp_failure}, we also observed some failure cases. Noted, many shadow removal approaches struggle with such scenes because the shadows occupy relatively large areas of the image and are indistinguishable from the image content.

\begin{table}[h]
\centering
\begin{tabular}{c|c}
\toprule
Network configurations                   & Hyper-parameter           \\
\midrule
input size                                & 64                                    \\
input channels                             & 11                                        \\
base channels                              & 128                                        \\
channel multipliers                        &{1, 1, 2, 2, 4, 4}                         \\
residual blocks per resolution             & 2                                         \\
attention resolution                       & 16 $\times$ 16                             \\
\bottomrule
\end{tabular}
\caption{Network configurations and parameter choices.}
\label{table:Network}
\end{table}

\begin{table}[t]
\centering
\begin{tabular}{c|c}
\toprule
Diffusion configurations                   & Hyper-parameter           \\
\midrule
diffusion steps ($T$) & 1000                                    \\
sampling timesteps ($S$)  & 25                                         \\
noise schedule ($\beta_t$)       & linear: 0.0001 $\rightarrow$ 0.02                                     \\
\bottomrule
\end{tabular}
\caption{Diffusion configurations and parameter choices.}
\label{table:Diffusion}
\end{table}

\begin{table}[!ht]
\centering
\resizebox{1.0\linewidth}{!}{
\begin{tabular}{c|ccc|ccc}
\toprule
 Resolution     & \multicolumn{3}{c|}{ShadowDiffusion} & \multicolumn{3}{c}{Ours} \\ 
  $256 \times 256$    
     & PSNR $\uparrow$       & SSIM $\uparrow$       & RMSE $\downarrow$       & PSNR $\uparrow$       & SSIM $\uparrow$       & RMSE $\downarrow$ \\
\midrule
  SRD & 30.67        & 0.910         & 4.21         &  \textbf{31.27}      &   \textbf{0.934}    & \textbf{3.61}  \\
\bottomrule
\end{tabular}
}
\caption{RMSE values on SRD dataset. 
}
\label{table:comp_with shadowdiffusion}
\end{table}

\begin{table}[t]
\centering
{
\begin{tabular}{l|cc|c}
\toprule
                            &RCA                                       & CA                  & PSNR$\uparrow$                             \\
\midrule
(a)                         &first-layer                               &w/o                    & 35.31                                      \\
(b)                         &w/o                                       &mid-layer              & 35.49                                      \\
(c)                         &second-layer                              &mid-layer              & 36.27                                      \\
(d)                         &third-layer                               &mid-layer              & 35.93                                      \\
(e)                         &first-layer\&mid-layer                    &w/o                    & 35.40                                      \\
(f)                         &w/o                                       &every-layer            & 36.14                                      \\
(h)                         &first-layer                              &mid-layer              & \textbf{36.47}                              \\
\bottomrule
\end{tabular}
}
\caption{Ablation studies to evaluate the effectiveness of different network structures on the ISTD+~\cite{le2019shadow} dataset.}
\label{table:more_ablation}
\end{table}

\noindent\textbf{Effectiveness of RCA module.}
We assess the impact of the RCA module and the Convolutional Aggregation (CA) 
at different layers on the ISTD+~\cite{le2019shadow} dataset. 
Table~\ref{table:more_ablation} (a) and (b) 
demonstrate the lower performance when only the RCA module or only the CA module is applied. Table~\ref{table:more_ablation} (c) highlights the importance of the RCA module, showing its effectiveness in shallow feature fusion because low-level features capture essential elements, such as edges and colors,
which are crucial for shadow removal tasks. Table~\ref{table:more_ablation} (d) confirms that adding the RCA module to both shallow and intermediate layers leads to unsatisfactory results due to limited mask size. Additionally, intermediate layer features primarily capture global semantic information, making them more suitable for fusion using the CA module. Table~\ref{table:more_ablation} (e) verifies the overall inferiority of solely incorporating the CA module at each layer compared to the full model in Table~\ref{table:more_ablation} (f) with the added RCA module. Therefore, the optimal approach is to apply the RCA module in the initial layer and the CA module in the intermediate layer.

\begin{figure}[t]
    \centering
    \includegraphics[width=1.0\linewidth]{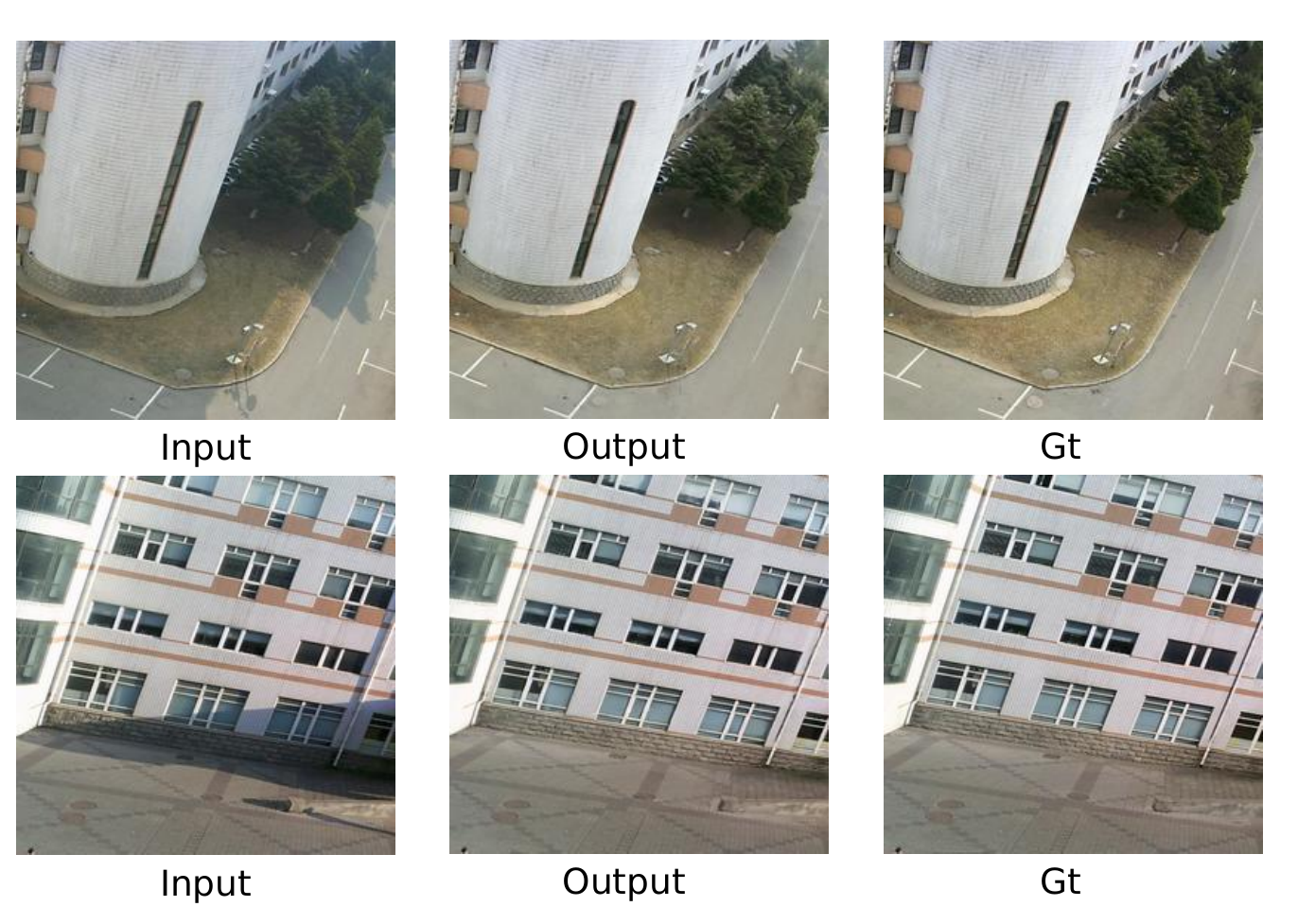}
    \caption{Some visual results on complex scenes.}
    \label{fig:complex_scene}
\end{figure}

\label{sec:visual}

\begin{figure*}[h]
    \centering
    \setlength{\belowcaptionskip}{2cm}
    \includegraphics[width=1\linewidth]{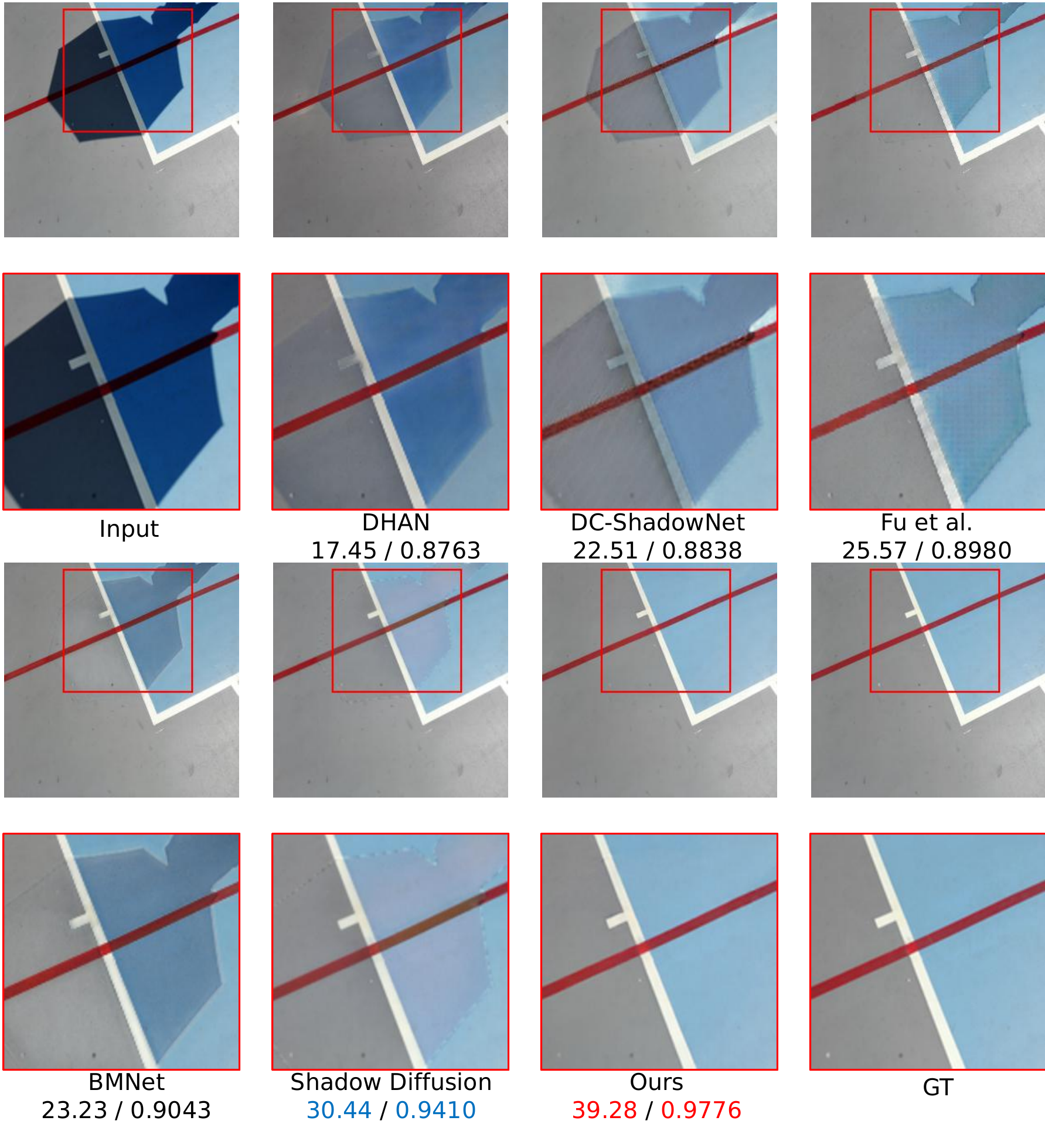} 
    \caption{One example of shadow removal results on the ISTD+~\cite{le2019shadow} dataset. Comparison with DHAN~\cite{cun2020towards}, DC-ShadowNet~\cite{jin2021dc}, Fu~\emph{et al.}~\cite{fu2021auto}, BMNet~\cite{zhu2022bijective} and ShadowDiffusion~\cite{guo2023shadowdiffusion}. 
    Corresponding values represent the `PSNR/SSIM' scores.
    The best and second-best results are illustrated in \textcolor{red}{red} and \textcolor{blue}{blue}.}
    \label{fig:comp_s1}
\end{figure*}
\begin{figure*}[ht!]
    \centering
    \setlength{\belowcaptionskip}{2cm}
    \includegraphics[width=1\linewidth]{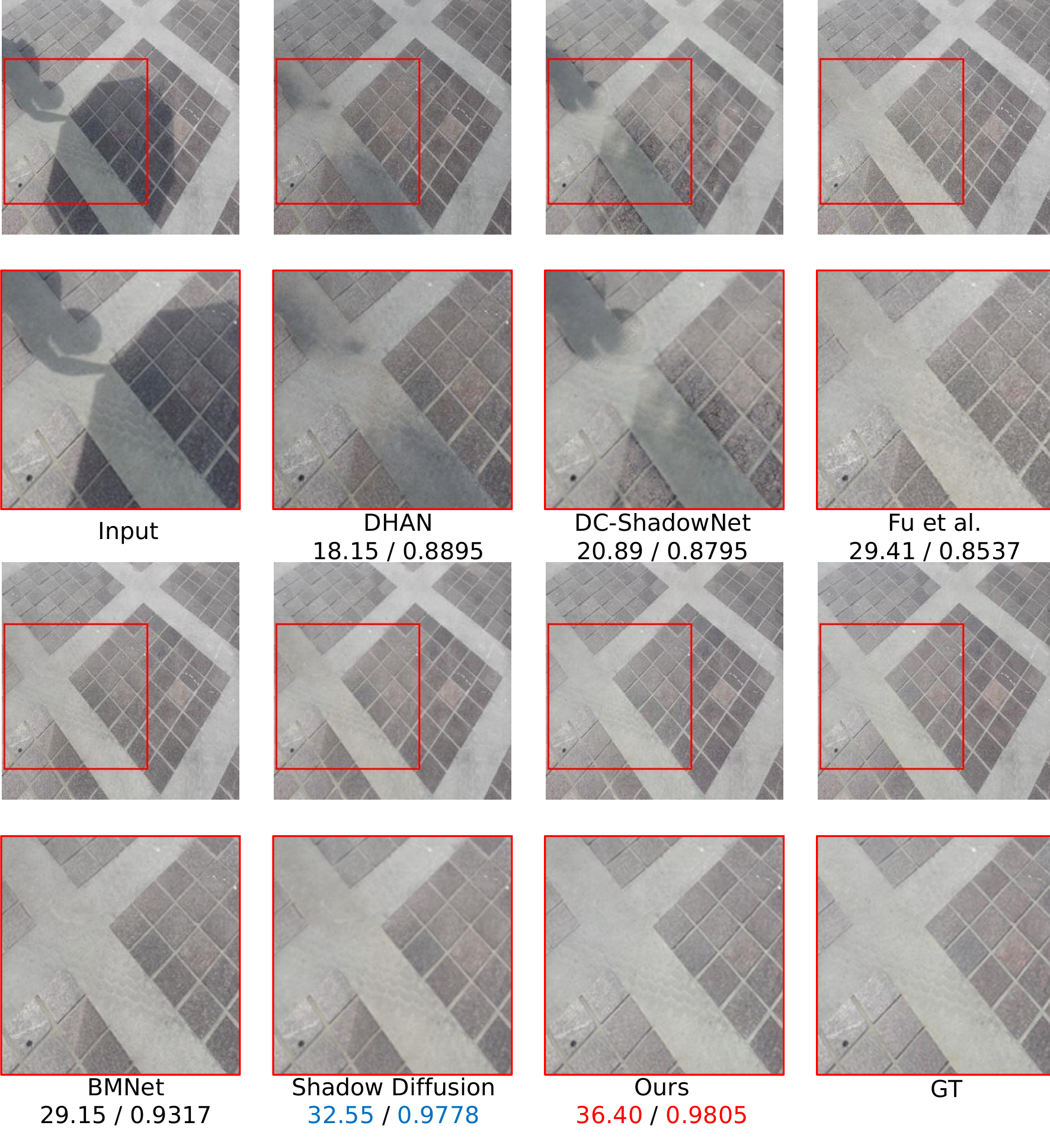} 
    \caption{One example of shadow removal results on the ISTD+~\cite{le2019shadow} dataset. Comparison with DHAN~\cite{cun2020towards}, DC-ShadowNet~\cite{jin2021dc}, Fu~\emph{et al.}~\cite{fu2021auto}, BMNet~\cite{zhu2022bijective} and ShadowDiffusion~\cite{guo2023shadowdiffusion} on ISTD+~\cite{le2019shadow} dataset. 
    Corresponding values represent the `PSNR/SSIM' scores.
    The best and second-best results are illustrated in \textcolor{red}{red} and \textcolor{blue}{blue}.}
    \label{fig:comp_s2}
\end{figure*}
\begin{figure*}[ht!]
    \centering
    \setlength{\belowcaptionskip}{2cm}
    \includegraphics[width=1\linewidth]{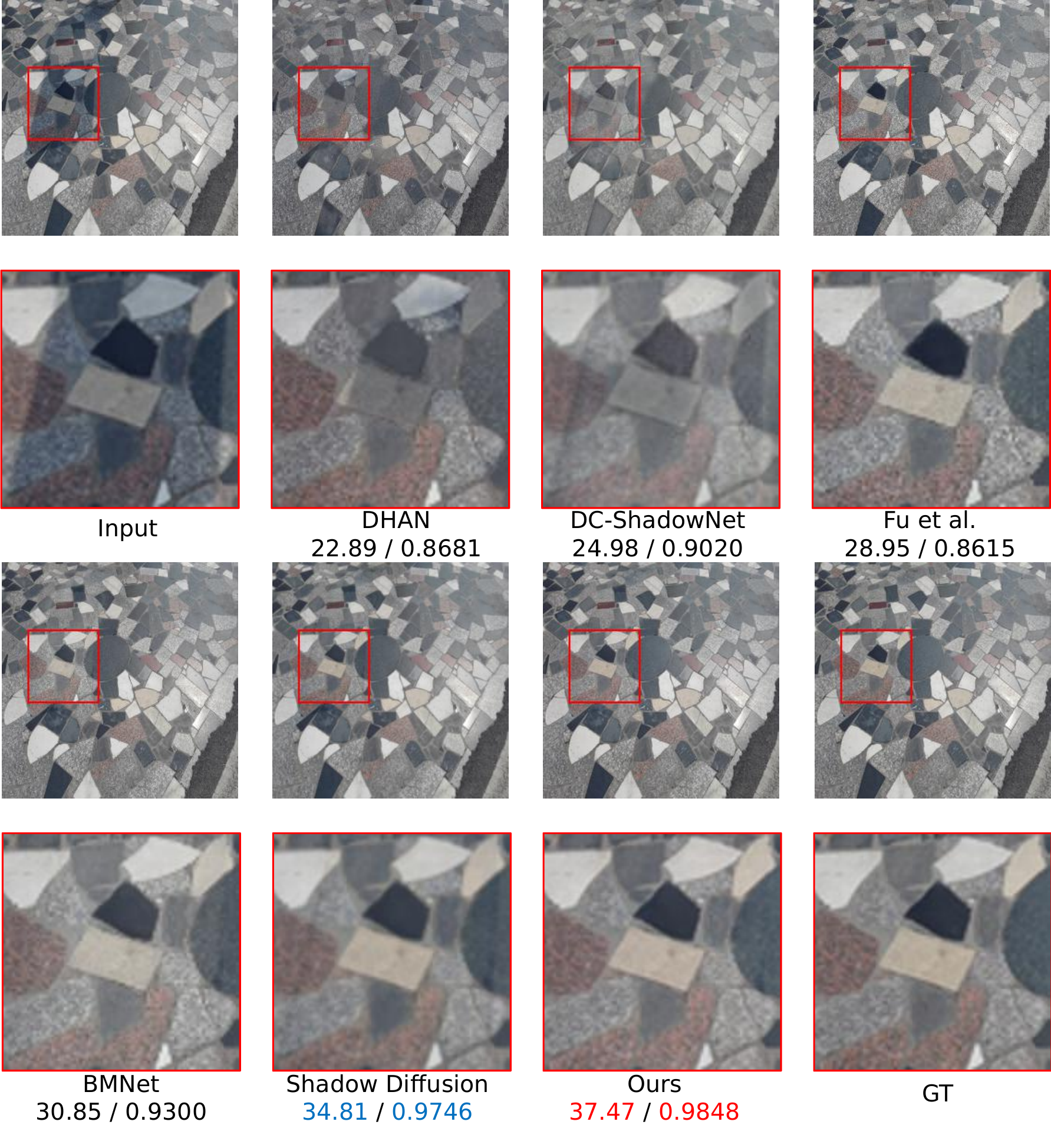} 
    \caption{One example of shadow removal results on the ISTD+~\cite{le2019shadow} dataset. Comparison with DHAN~\cite{cun2020towards}, DC-ShadowNet~\cite{jin2021dc}, Fu~\emph{et al.}~\cite{fu2021auto}, BMNet~\cite{zhu2022bijective} and ShadowDiffusion~\cite{guo2023shadowdiffusion}. 
   Corresponding values represent the `PSNR/SSIM' scores.
    The best and second-best results are illustrated in \textcolor{red}{red} and \textcolor{blue}{blue}.}
    \label{fig:comp_s3}
\end{figure*}
\begin{figure*}[ht!]
\setlength{\belowcaptionskip}{3cm}
    \centering
    \includegraphics[width=1\linewidth]{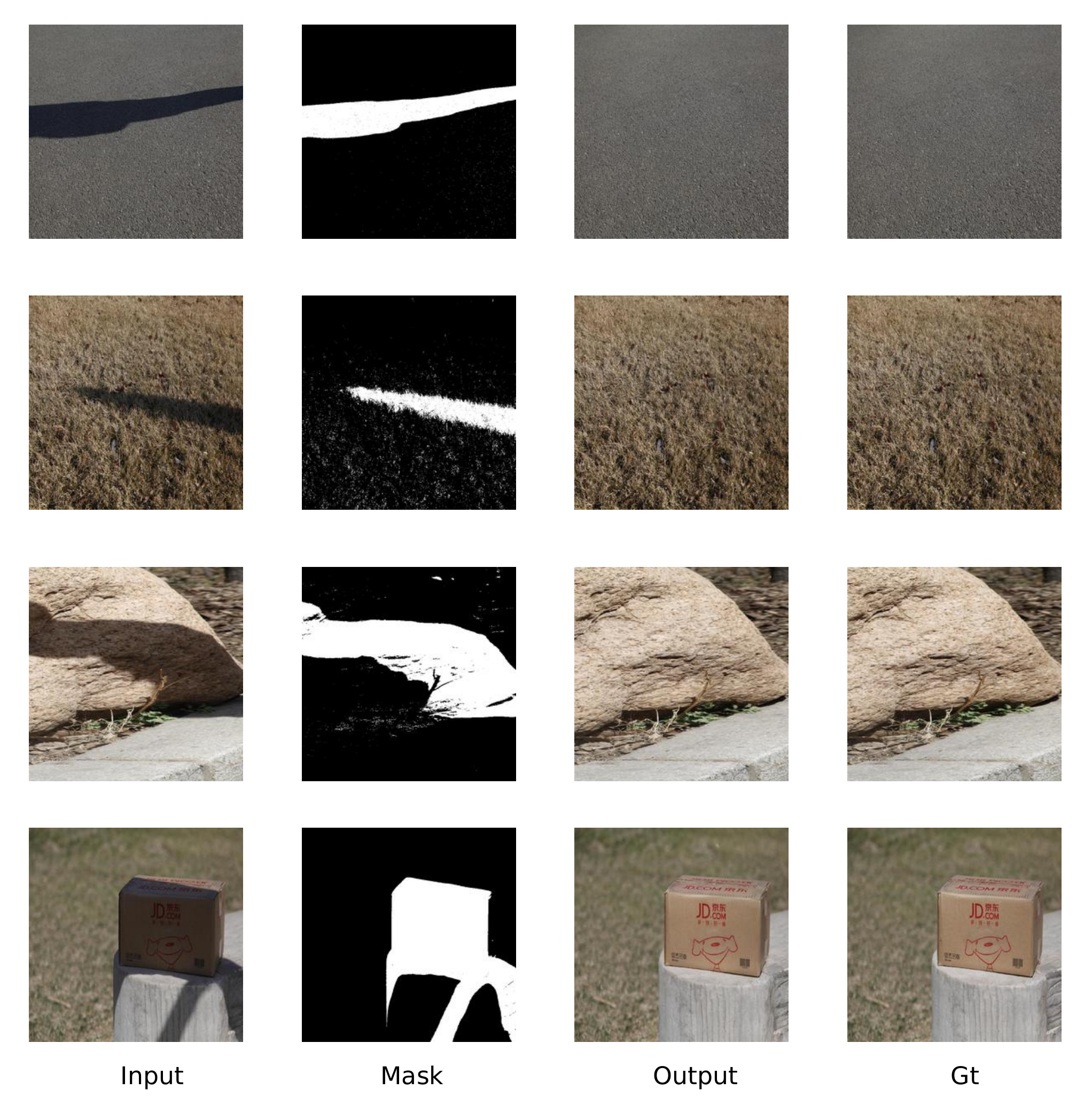} 
    \caption{Shadow removal results of Diff-Shadow on the SRD~\cite{qu2017deshadownet} dataset. }
    \label{fig:res_s1}
\end{figure*}
\begin{figure*}[ht!]
\setlength{\belowcaptionskip}{3cm}
    \centering
    \includegraphics[width=1\linewidth]{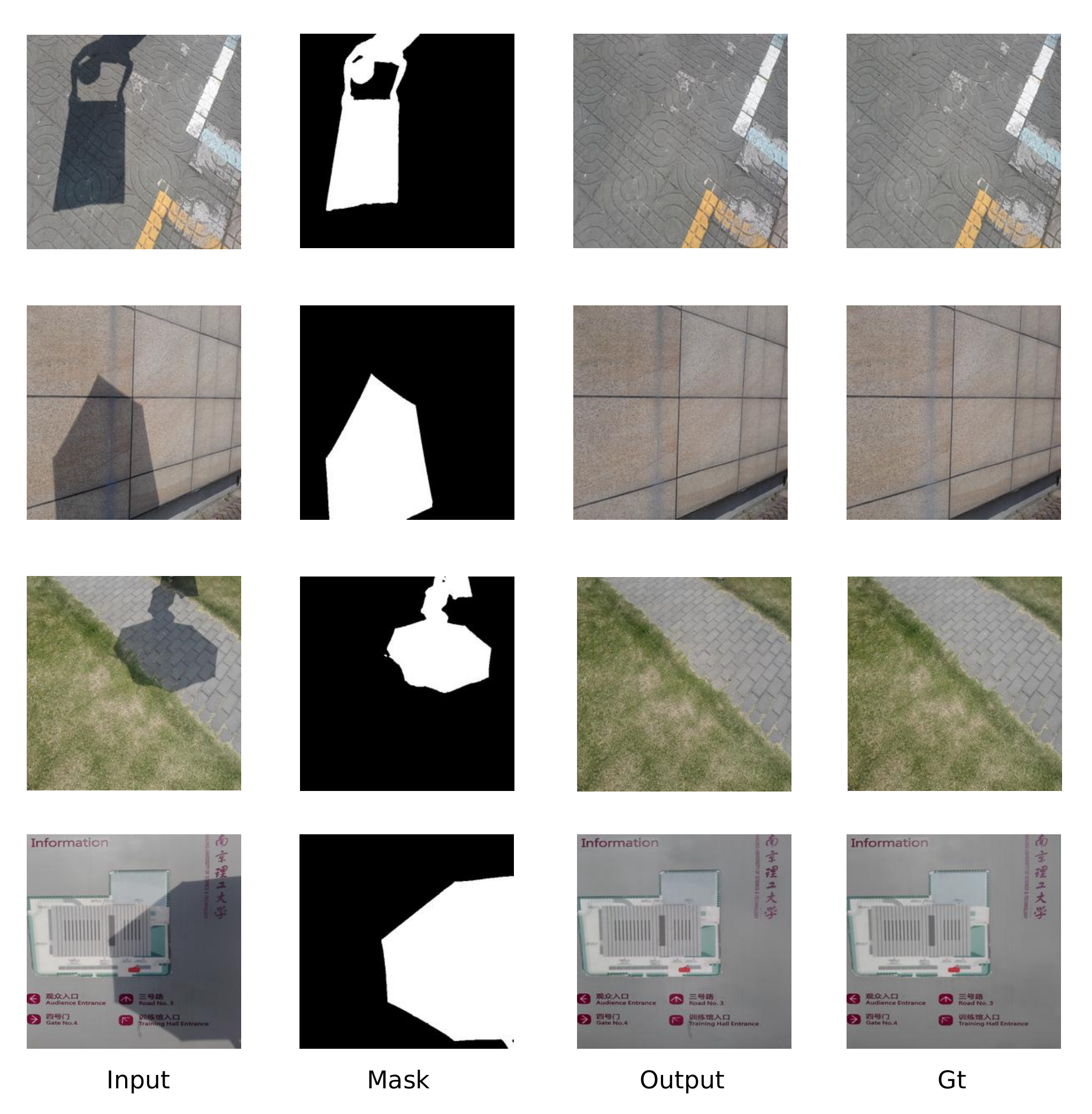} 
    \caption{Shadow removal results of Diff-Shadow on the ISTD~\cite{wang2018stacked} dataset. }
    \label{fig:res_s2}
\end{figure*}

\noindent\textbf{More Visual Results.}
Figure~\ref{fig:complex_scene} presents some visual results on complex scenes to demonstrate the robustness of the proposed method.
Figures~\ref{fig:comp_s1},~\ref{fig:comp_s2},~\ref{fig:comp_s3} illustrate some qualitative comparison with DHAN~\cite{cun2020towards}, DC-ShadowNet~\cite{jin2021dc}, Fu~\emph{et al.}~\cite{fu2021auto}, BMNet~\cite{zhu2022bijective}, and ShadowDiffusion~\cite{guo2023shadowdiffusion} on ISTD+~\cite{le2019shadow} dataset.  Our approach clearly outperforms previous methods by preserving the image texture and the luminance consistency. Figure~\ref{fig:res_s1} and Figure~\ref{fig:res_s2} exhibit some results of the proposed Diff-Shadow on SRD~\cite{qu2017deshadownet} and ISTD~\cite{wang2018stacked} datasets, respectively. 

\begin{figure}[t]
    \centering
    \includegraphics[width=1.0\linewidth]{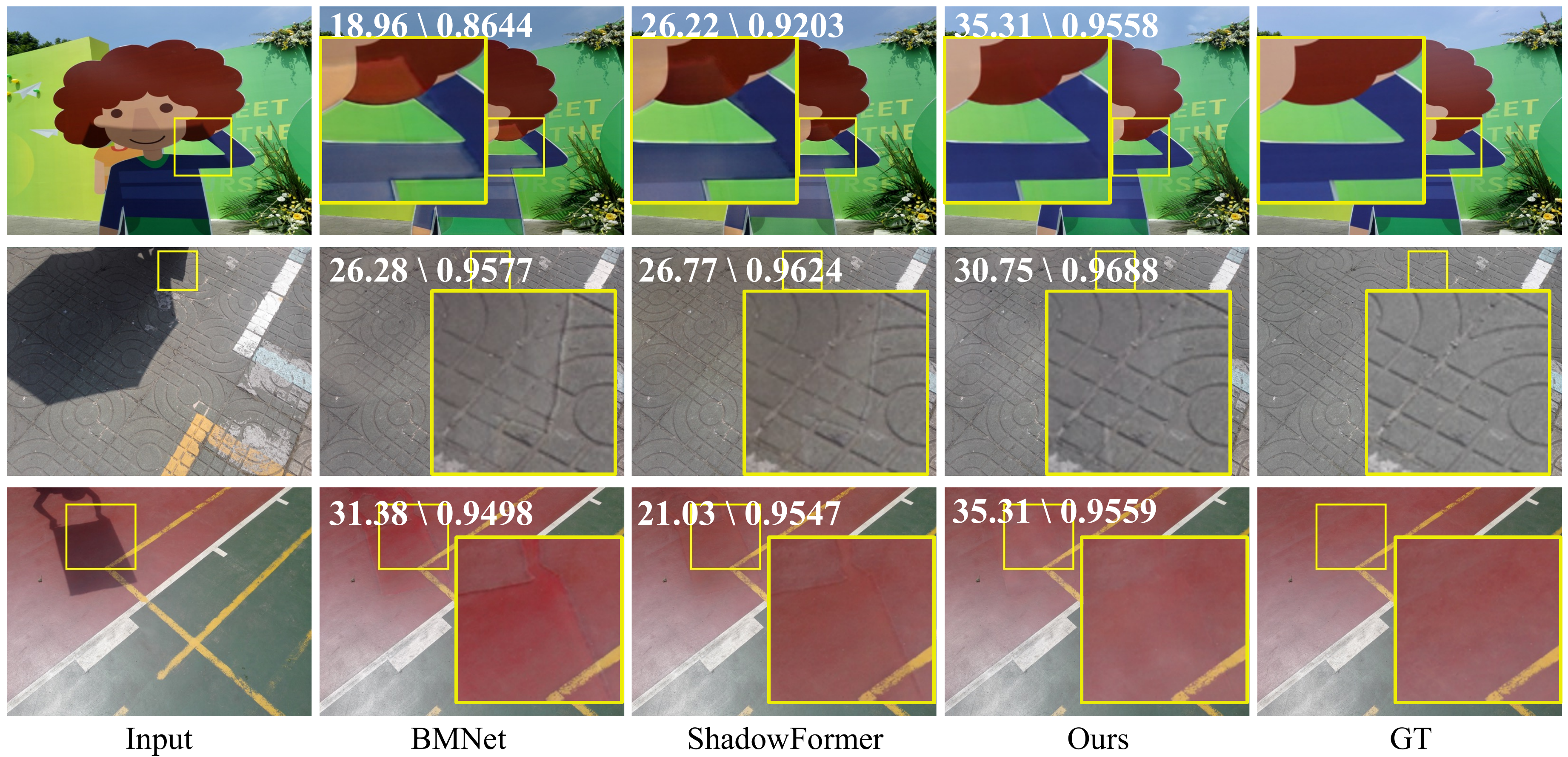}
    \caption{Qualitative comparison with original resolution for BMNet, ShadowFormer, and ours on ISTD~\cite{le2019shadow} dataset. }
    \label{fig:comp-full}
\end{figure}

\begin{table}[!ht]
\centering
\resizebox{1.0\linewidth}{!}{
\begin{tabular}{c|c|ccc}
\toprule
Resolution               & Method                                                  & PSNR$\uparrow$                  & SSIM$\uparrow$                  & RMSE$\downarrow$     \\  
\midrule
\multirow{5}{*}{\begin{tabular}[c]{@{}c@{}}original size\\ $640 \times 480$  \end{tabular}} & Input Image     & 20.33                           & 0.874                           & 11.35    \\
   & DHAN~\cite{cun2020towards}              & 27.88                           & 0.921                           & 6.29    \\
   & BMNet~\cite{zhu2022bijective}           & 29.02                           & 0.928                           & 4.17   \\
   & ShadowFormer~\cite{guo2023shadowformer} & 30.47                           & 0.935                           & 4.79  \\
   & \textbf{Ours}             &\textbf{30.58}                  &\textbf{0.938}                    & \textbf{3.69}   \\ 
\bottomrule
\end{tabular}
}
\caption{Quantitative results on images with original resolution from ISTD~\cite{wang2018stacked} dataset.}
\label{table:original}
\end{table}

\noindent\textbf{Extension to Other Image Resolutions.}
Thanks to the patch-based diffusion modeling approach and the sampling process across overlapping patches during inference, the proposed method enables size-agnostic image restoration. As shown in Fig.~\ref{fig:comp-full}, our Diff-Shadow maintains excellent illumination consistency even at the original image resolution ($640 \times 480$) of the ISTD~\cite{le2019shadow} dataset. Furthermore, as illustrated in Table~\ref{table:original}, the proposed Diff-Shadow also achieves state-of-the-art performances. Note that the comparison methods are trained on images of original resolution, while our training is based on images with a resolution of $256 \times 256$.

These results demonstrate the remarkable effectiveness of our approach in removing shadows across diverse scenes due to the incorporation of global information guidance and the powerful modeling capability of the diffusion model for image distribution.

\section{Differences with Other Methods}
\noindent\textbf{Differences with WeatherDiffusion.}
The main differences include two aspects. 
First, in terms of network architecture, in contrast to WeatherDiffsuion's~\cite{ozdenizci2023restoring} simple single-head UNet architecture, we adopt the parallel UNets with a local branch dedicated to noise estimation and a global branch designed for shadow-free image recovery. Additionally, we introduce the RCA module to
avoid the problem of losing global information for patch recovery in WeatherDiffsuion. Second, regarding the sampling process, compared to the average weighting used in WeatherDiffsuion, our global guided sampling strategy (GSS) adaptively calculates patch fusion weights, ensuring the consistency of illumination in the shadow removal process. Finally, the ablation studies have demonstrated the considerable improvements.

\noindent\textbf{Differences with CRFormer.}
First, in contrast to cross-attention modules in CRFormer~\cite{wan2022crformer} whose $\mathbf{Q}$, $\mathbf{K}$ and $\mathbf{V}$ originates from the same global feature, the RCA integrates both local and global information. 
Furthermore, other attention-based methods boost the mutual information between shadow and non-shadow pairs through masks. 
However, the shadow information is dispensable for recovering non-shadow regions.
Contrarily, our attention mechanism utilizes masks to suppress the global shadow values,
integrating the non-shadow information from the global branch into the local branch. Finally, Table \ref{table:re_comp} shows the performance improvements compared with CRFormer.

\begin{table}[h]
\centering
\begin{tabular}{c|ccc|ccc}
\toprule
      & \multicolumn{3}{c|}{CRFormer} & \multicolumn{3}{c}{Ours} \\ 
      
      & S       & NS       & ALL       & S       & NS       & ALL \\
\midrule
ISTD+ &  5.90       &  2.90        &    3.40       & \textbf{1.30}      &    \textbf{1.57}    &   \textbf{2.09}     \\
SRD   &  7.14       &  3.15        &   4.25       &  \textbf{1.81}      &   \textbf{1.65}    & \textbf{2.52}  \\
\bottomrule
\end{tabular}
\caption{RMSE values on ISTD+ and SRD datasets. 
}
\label{table:re_comp}
\end{table}

\section{Limitations}
\label{sec:limitations}
The main limitation of our approach is the relatively long inference time compared to end-to-end image recovery networks and diffusion methods using simple UNet architectures. Our Diff-Shadow takes 3.5 seconds (grid = 16) to recover an image of size $256 \times 256$ on a single NVIDIA GTX 2080Ti GPU. The number of parameters in our method grows from 104M to 160M compared to the single-branch architecture, we believe that the increase in computation actually correlate well with improvements in results.

Another natural limitation of our model is the requirement for accurate shadow masks. This enables the network to efficiently use information from the non-shaded portion of the global image, and thus the accuracy of shadow detection directly determines the final result of the shadow removal. Similar to~\cite{guo2023shadowformer,guo2023shadowdiffusion,hu2019direction}, we apply the masks from  ~\cite{cun2020towards} for fair comparisons, which is widely used on shadow removal tasks.
\end{document}